%% file: main.tex
\documentclass[10pt, a4paper]{article}

\usepackage[utf8]{inputenc}
\usepackage[T1]{fontenc}
\usepackage[english]{babel}
\usepackage{geometry}
\usepackage{booktabs} 
\usepackage{graphicx} 
\usepackage{xcolor}
\usepackage{fancyhdr}
\usepackage{authblk} 
\usepackage{natbib}
\usepackage{booktabs}
\usepackage{CJKutf8}
\usepackage[utf8]{inputenc}
\usepackage[most]{tcolorbox}

\usepackage{placeins}
\usepackage{float}
\usepackage{times}
\usepackage{latexsym}
\usepackage{amsmath}
\usepackage{amsfonts}
\usepackage{multirow}
\usepackage{subcaption}
\usepackage{tabularx}

\usepackage{booktabs}
\usepackage{multirow}
\usepackage{siunitx}
\usepackage{enumitem}

\usepackage{amssymb}

\usepackage[T1]{fontenc}

\usepackage[utf8]{inputenc}

\usepackage{microtype}

\usepackage{inconsolata}

\usepackage{graphicx}

\definecolor{bilipink}{RGB}{251, 114, 153}
\definecolor{biliblue}{RGB}{0, 140, 210}

\usepackage{listings}
\usepackage[ruled,vlined]{algorithm2e}
\definecolor{commentcolor}{RGB}{110,110,110}

\SetCommentSty{mycommfont}

\usepackage[default]{sourcesanspro}

\usepackage[
    colorlinks,
    citecolor=bilipink,
    linkcolor=bilipink,
    urlcolor=bilipink,
    breaklinks=true
]{hyperref}
\fancypagestyle{firstpage}{
    \fancyhead[R]{
        \makebox[0pt][r]{
            \raisebox{-0.42\height}{\includegraphics[height=1.6cm]{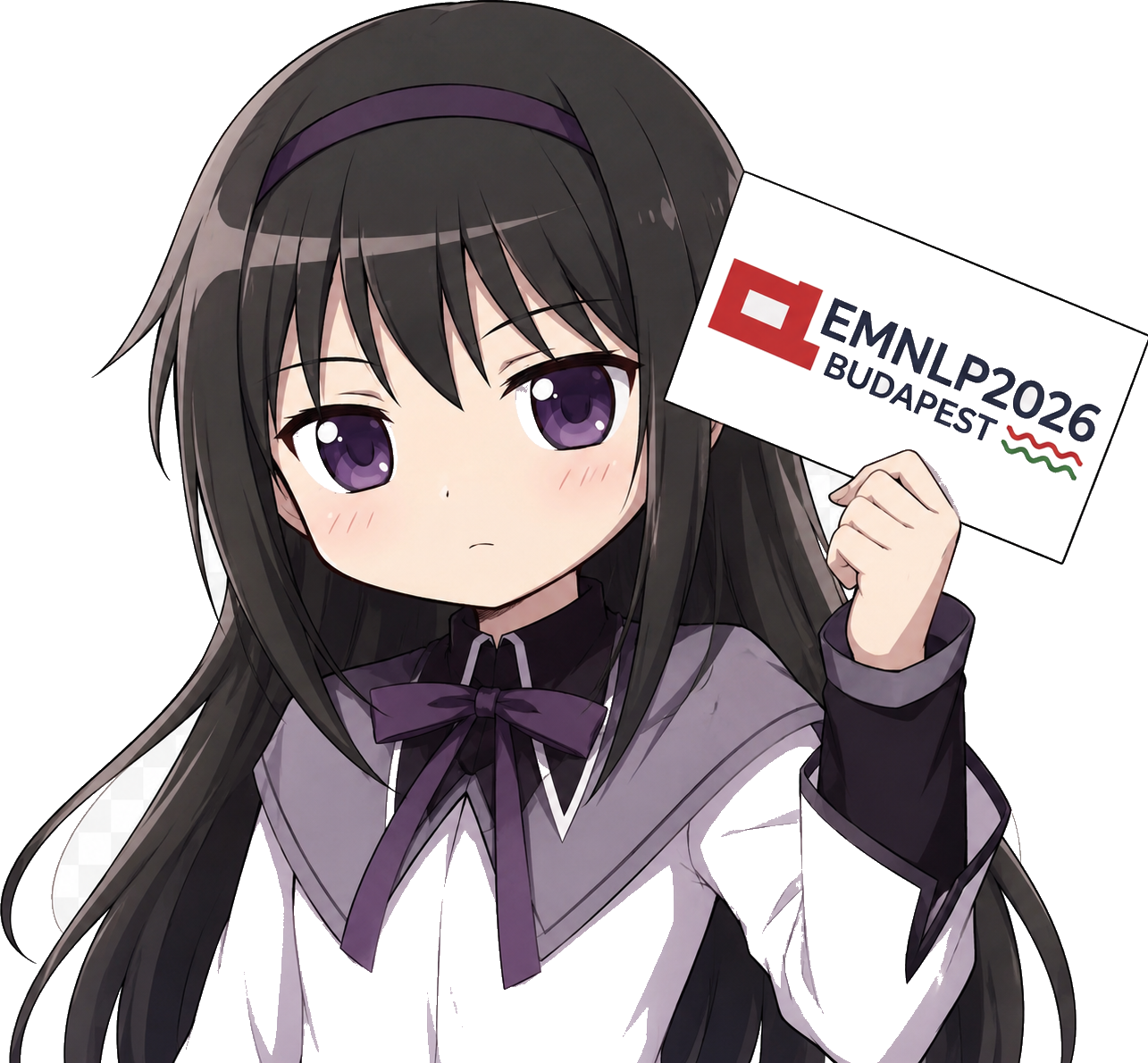}}
            \hspace{-0.3cm}
        }
    }
}

\makeatletter
\def\@maketitle{%
  \newpage
  \null
  \vspace{-3.5em} 
  \begin{center}%
    {\color{biliblue}\hrule height 0.8pt}
    \vspace{1.5em}
    
    {\huge \bfseries \@title \par} 
    \vspace{1.5em}
    
    {\color{biliblue}\hrule height 0.8pt}
    \vspace{1.8em}
    
    {\large \bfseries \@author} \\
    \vspace{0.5em}
  \end{center}%
  \par
  \vspace{1.5em}}
\makeatother

\usepackage{hyperref}
\hypersetup{
    colorlinks=true,
    urlcolor=bilipink, 
}

\title{\textsc{Homura}: Taming the Sand-Glass for Time-Constrained LLM Translation via Reinforcement Learning}
\author{
    {\normalsize
    \ \ \textbf{Ziang Cui}$^{*}$,
     \ \ \textbf{Mengran Yu}$^{*}$,
     \ \ \textbf{Chenyu Shi},
     \ \ \textbf{Yingxuan Shi},
     \ \ \textbf{Tianjiao Li}$^{*, \dagger}$
     }\\
    \vspace{4pt}
    \small Bilibili Inc., Shanghai, China \\
    \vspace{2pt}
    \small $^*$Equal contribution \quad $^\dagger$Corresponding author \\
    \vspace{2pt}
    \texttt{\{cuiziang, yumengran03, shichenyu, shiyingxuan, litianjiao01\}@bilibili.com}
}
\date{} 

\begin{document}
\begin{CJK*}{UTF8}{gbsn} 
\maketitle
\thispagestyle{firstpage} 

\begin{abstract}
\noindent Large Language Models (LLMs) have achieved remarkable strides in multilingual translation but are hindered by a systemic cross-lingual verbosity bias, rendering them unsuitable for strict time-constrained tasks like subtitling and dubbing. Current prompt-engineering approaches struggle to resolve this conflict between semantic fidelity and rigid temporal feasibility. To bridge this gap, we first introduce Sand-Glass, a benchmark specifically designed to evaluate translation under syllable-level duration constraints. Furthermore, we propose \textsc{Homura}, a reinforcement learning framework that explicitly optimizes the trade-off between semantic preservation and temporal compliance. By employing a constrained reinforcement learning objective featuring a novel dynamic syllable-ratio reward, \textsc{Homura} effectively "tames" the output length. Experimental results demonstrate that \textsc{Homura} significantly outperforms strong baselines, achieving precise length control that respects linguistic density hierarchies without compromising semantic adequacy.\footnote{Our code and data are available at: \url{https://github.com/InvokerStark/HOMURA}.}
\end{abstract}

\input{section/IntroductionV2}
\input{section/Pilot_Experiment_Analysis}

\input{section/Method}

\input{section/Experiment}

\input{section/Conclusion}

\section*{Limitations}

First, while we adopt syllable count as a duration proxy grounded in the Iso-Information Principle, this metric remains a textual approximation of acoustic reality. In real-world dubbing and subtitling, physical duration is heavily influenced by prosodic features and speech tempo, and our current text-only framework does not account for multimodal constraints like lip-synchronization or isochrony. Consequently, generated outputs may still require minor timing adjustments during recording.

Second, our experimental validation is currently restricted to Chinese-to-English, German, and Spanish translation tasks. The universality of the "compression feasibility boundary" observed in our study remains to be verified across linguistically distant pairs or low-resource languages. Future work is needed to determine if such limits are language-specific or intrinsic to the translation task itself.

\bibliographystyle{plainnat} 
\bibliography{refs}      

\appendix
\input{section/Appendix}
\end{CJK*} 
\end{document}

%% file: section/IntroductionV2.tex
\section{Introduction}

\begin{figure}[t]
\centering
  \includegraphics[width=\linewidth]{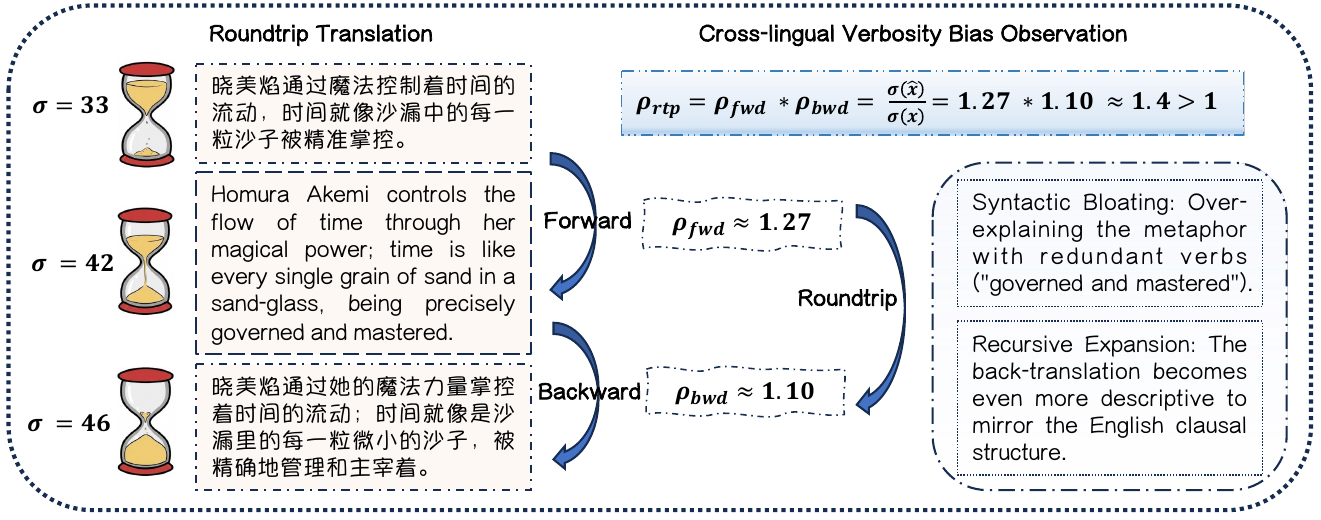}
\caption{Illustration of cross-lingual verbosity bias. Unlike forward metrics, the \textbf{Roundtrip Expansion Ratio} $\rho_{\text{rtp}}$ isolates model-induced redundancy from linguistic density shifts. A value of $\rho_{\text{rtp}} > 1.0$ indicates systemic inflation despite constant semantic content. Formal definition and diagnostic study are provided in Section~\ref{sec:pilot}.}
 \label{fig:roundtrip_case}
\end{figure}

With the rapid advancement of Large Language Models (LLMs), their multilingual capabilities have achieved unprecedented semantic adequacy and fluency in Neural Machine Translation (NMT)~\cite{hendy2023good, jiao2023chatgpt, zhu2024multilingual}. However, this linguistic prowess is accompanied by a systemic cross-lingual verbosity bias~\cite{briakou2024implications, manakhimova2024investigating}. As illustrated in Figure~\ref{fig:roundtrip_case}, LLM-generated translations tend to be significantly longer than source utterances, a "chattiness" that persists even when semantic content remains constant. While acceptable in general text generation, this expansion fails in time-constrained multimedia localization (e.g., subtitles, dubbing, interpretation), where translation length constitutes a rigid temporal budget that strictly limits output duration~\cite{karakanta-etal-2020-must, jorgensen-mengshoel-2025-cross, gaido2024automatic}.
To operationalize the temporal budget, we approximate the available speaking time with a temporally grounded syllable budget scaled to each target language.

Viewed through an information-theoretic lens ~\cite{shannon1948mathematical, davisson1972rate}, these temporal budgets act as a strict rate constraint: with a limited number of target syllables available, the model must pack maximal meaning into minimal length, operating near a rate-distortion limit on the achievable fidelity compression trade-off. In our setting, the rate is the syllable budget, and distortion can be understood as the loss in meaning. 

Unfortunately, off-the-shelf LLMs struggle to respect these limits. This failure is rooted in the post-training alignment process, where learned reward models frequently exhibit a "length bias" that conflates verbosity with quality~\cite{saito2023verbosity, singhal2023long}. 
Moreover, cross-lingual typological differences systematically skew length ratios, transforming length control from a formatting task into a first-class modeling challenge.
\footnote{Detailed related works can be found in appendix \ref{sec:appendix_related_works}.}

To address this fundamental mismatch between LLM capabilities and temporal constraints, we argue that length control should be treated as an optimization objective rather than a post-hoc adjustment. To this end, we first operationalize the problem by constructing \emph{Sand-Glass}, a dedicated benchmark for time-constrained media localization. The name reflects the rigid temporal budget that "drains" as the model produces redundant tokens. 
Unlike generic translation datasets, Sand-Glass incorporates syllable-based duration proxies calibrated by Information Density (ID), derived from real-world colloquial subtitles enabling precise evaluation of length compliance.

Building on this testbed, we propose \textsc{Homura} (\textbf{H}ard \textbf{O}ptimization for \textbf{M}ultilingual \textbf{U}tterance \textbf{R}eduction and \textbf{A}lignment), a reinforcement learning framework designed to ``tame the sandglass'' by explicitly regulating translation length without requiring supervised compression datasets. Instead of relying on brittle prompts, \textsc{Homura} optimizes the quality compression trade-off directly. We design a novel reference-free reward function that combines a dynamic syllable ratio penalty with a semantic fidelity reward, encouraging the model to condense information density without sacrificing core meaning. Our core contributions are as follows:
\begin{itemize}[topsep=2pt, itemsep=0pt, parsep=0pt, leftmargin=*]
    \item \textbf{Diagnostic Analysis:} We systematically quantify the verbosity bias in LLM-based translation under temporal constraints, providing cross-lingual diagnostics (e.g., the $\rho_{\text{rtp}}$ metric) that reveal the limitations of unconstrained models as visualized in Figure~\ref{fig:roundtrip_case}.
    \item \textbf{Benchmark Construction:} We introduce \textsc{Sand-Glass}, a specialized dataset tailored for time-constrained translation, featuring syllable-level duration budgets to simulate real-world dubbing and subtitling scenarios.
    \item \textbf{Methodological Innovation:} We introduce \textsc{Homura}, a constrained RL framework for temporally grounded translation, which substantially improves budget compliance while maintaining semantic fidelity over strong LLM baselines.
\end{itemize}

%% file: section/Pilot_Experiment_Analysis.tex
\section{Pilot Study: Benchmark and Verbosity Bias}
\label{sec:pilot}

\begin{figure*}[th]
\centering
\includegraphics[width=0.95\textwidth]{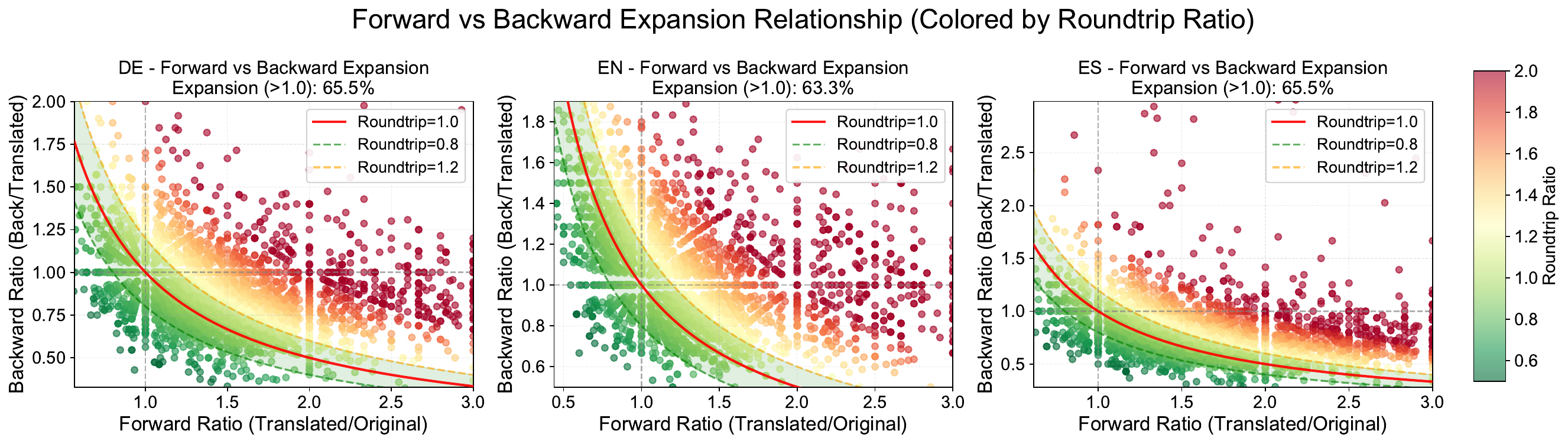}
\vspace{-2mm}
\caption{Forward vs.\ backward expansion relationship (colored by round-trip ratio $\rho_{\text{rtp}}$). The red curve represents the unity baseline $\rho_{\text{rtp}}=1.0$ ($y=1/x$). Points above this curve indicate systemic model-induced length inflation. Unconstrained models show a median $\rho_{\text{rtp}} > 1.10$.}
\label{fig:scatter_forward_backward}
\end{figure*}

\subsection{The Sand-Glass Benchmark}

To diagnose cross-lingual verbosity bias under temporal constraints, we introduce \textsc{Sand-Glass}, a benchmark built from colloquial video transcripts. We collect real-world videos from five domains---\textit{Gaming}, \textit{Film \& Television}, \textit{Travel \& Tourism}, \textit{ACGN}, and \textit{General Knowledge}---and segment ASR transcripts into subtitle-style units with explicit duration budgets. 


\paragraph{Construction Procedure.}
\begin{enumerate}[leftmargin=*, label=\textbf{(\arabic*)}, itemsep=0.6em, parsep=0pt]
    \item \textbf{Data Collection and Segmentation.}
    We transcribe real-world videos with ASR and segment the transcripts according to pauses and semantic boundaries. Each segment inherits its original speech duration as the temporal budget for translation.

    \item \textbf{Preprocessing and Packaging.}
    We remove non-speech artifacts such as fillers and music markers, and package segments into ten-unit sliding windows to preserve local context. All instances are formatted into a fixed JSON schema for model interaction.

    \item \textbf{Filtering and Sampling.}
    We filter out overly short or extreme-CPS segments to ensure stable syllable-ratio estimation, and further score samples using heuristic signals such as perplexity, repetition, and script consistency. After LSH-based de-duplication~\cite{jha2023limit}, we perform domain-balanced quota sampling to obtain \textit{1,000 golden instances}.

    \item \textbf{Core Event Annotation.}
    We extract golden core events from each source Mandarin segment by identifying content-bearing predicate--argument structures (nouns, proper nouns, numerals) via Stanza.\footnote{\url{https://stanfordnlp.github.io/stanza/}} To evaluate semantic preservation under length compression, candidate translations are first back-translated into Mandarin. We then extract candidate event realizations using the same pipeline. Source and back-translated event representations are compared via contextual embedding similarity; unmatched events are flagged as semantic violations to compute \textit{BT-CERR}.
\end{enumerate}
Following this pipeline, the finalized \textsc{Sand-Glass} benchmark contains 1,000 high-quality golden instances balanced across five distinct media domains. Detailed descriptive statistics are deferred to Appendix~\ref{sec:appendix_dataset_stats}.

\subsection{Syllable as Duration Proxy}
\label{sec:syllable_proxy}
Our use of syllables as duration proxies is grounded in the Iso-Information Principle. \citet{Pellegrino2011language} demonstrated that while Syllable Rate (SR) vary drastically across languages (e.g., Spanish $7.82 \sigma/s$ [syllables/s] vs.\ Mandarin $5.18 \sigma/s$), the universal Information Rate remains nearly constant at $\approx 39$ bits/s~\cite{Coupe2019efficiency}.Recent findings further corroborate this by quantifying a universal surprisal–duration trade-off \citep{pimentel-etal-2021-surprisal}, showing that languages systematically adjust segment duration to smooth information transmission.

\begin{table}[t]
\centering
\small
\setlength{\tabcolsep}{3pt}
\begin{tabular}{lccc}
\toprule
\textbf{Language} & \textbf{Syllable Rate} & \textbf{Info. Density} & \textbf{Expansion} \\
& ($\sigma/s$) & (Normalized) & (from Zh) \\
\midrule
Mandarin (Zh) & $5.18 \pm 0.15$ & $0.94$ & $1.00\times$ \\
English (En) & $6.19 \pm 0.16$ & $0.91$ & $\approx 1.03\times$ \\
German (De) & $5.97 \pm 0.19$ & $0.79$ & $\approx 1.19\times$ \\
Spanish (Es) & $7.82 \pm 0.16$ & $0.63$ & $\approx 1.49\times$ \\
\bottomrule
\end{tabular}
\vspace{-2mm}
\caption{Cross-linguistic statistics adapted from \citet{Pellegrino2011language}. Lower density necessitates higher expansion factors when translating from Mandarin.}
\label{tab:density_stats}
\vspace{-3mm}
\end{table}

As shown in Table~\ref{tab:density_stats}, this trade-off implies that low-density languages must utilize higher syllable counts to convey equivalent content. Translating from a high-density source (Mandarin, ID=0.94) to a lower-density target (e.g., Spanish, ID=0.63) necessitates a \textit{linguistically required expansion}. A direct 1:1 mapping would result in $\approx 33\%$ information loss (derived from the density ratio $\frac{0.63}{0.94}$). Thus, our constraints are non-uniform and calibrated to these density ratios to distinguish legitimate expansion from model verbosity.

\textbf{Empirical validation.} To verify that syllable count reliably predicts real speech duration, we synthesize the benchmark's English translations with 9 TTS configurations across diverse synthesis paradigms and measure actual silence-trimmed durations. Syllable count correlates almost perfectly with measured duration across all systems ($r = 0.89\text{--}0.98$); adding a simple punctuation-break count (periods, commas, question marks) raises explained variance to $R^2 = 0.93\text{--}0.96$, leaving at most 5–7\% of variance to free prosody. Crucially, when we synthesize each baseline's output against the original time budgets, unconstrained LLMs run 32\% over the slot, prompt/pipeline baselines still exceed by over 10\%, and Best-of-$N$ fits only by undershooting ($0.87\times$). In contrast, \textsc{Homura}'s spoken duration is centered on the real budget (median ratio $= 1.02$) in a single pass. Thus, syllable count is an empirically validated first-order proxy for speaking time, and optimizing it delivers time-constrained behavior in actual speech (see Appendix~\ref{app:tts_validation} for full details).

\subsection{Quantifying Cross-Lingual Verbosity Bias}
\label{sec:verbosity_bias}

Standard metrics like the Forward Expansion Ratio defined as $\rho_{fwd} = \sigma(y)/\sigma(x)$ for a source $x$ and its translation $y$ often confound \textit{linguistic necessity} with \textit{model hallucinations}. For instance, a high $\rho_{fwd}$ in Spanish might simply reflect its naturally lower information density rather than model-induced redundancy. 

To isolate systemic bias, we introduce the \textit{Roundtrip Expansion Ratio} ($\rho_{rtp}$). By translating the target output $y$ back into the source language ($\hat{x}$) and comparing its syllable count to the original input $x$, we define:
\begin{equation}
    \rho_{rtp} = \rho_{fwd} \cdot \rho_{bwd} = \frac{\sigma(y)}{\sigma(x)} \cdot \frac{\sigma(\hat{x})}{\sigma(y)} = \frac{\sigma(\hat{x})}{\sigma(x)},
\end{equation}
where $\sigma(\cdot)$ denotes the syllable count of a sequence. Theoretically, since this comparison is performed within the same source language, $\rho_{rtp}$ should gravitate towards 1.0. Any value where $\rho_{rtp} > 1.0$ serves as a density-invariant indicator of model-induced inflation.

Empirical analysis on Sand-Glass (Figure~\ref{fig:scatter_forward_backward}) confirms that this verbosity is systemic rather than anecdotal. By plotting the joint distribution of expansion ratios, we observe a pervasive upward shift from the unity baseline ($\rho_{rtp}=1.0$), with the vast majority of segments clustered in the inflation zone. This concentration demonstrates a consistent model-induced redundancy that persists across various frontier models and language pairs (see Appendix~\ref{sec:appendix_noraml_roundtrip_metrics} for statistical breakdowns).

\paragraph{Calibrated Target Bounds.}
To bridge this gap, we define Target Syllable Bounds ($\mathcal{B}_{L}$) by contrasting model inflation ($\mu_{LLM}$) against Theoretical Expansion ($\mu_{theo}$). As shown in Table~\ref{tab:expansion_comparison}, LLMs produce significant surplus bias ($\Delta_{bias}$). We establish $\mathcal{B}_{L}$ to enforce a strict condensation regime. Notably, we lower these targets beyond theoretical baselines to address real-world dubbing and subtitling complexities, specifically to buffer against rapid speech and synchronization constraints. Accordingly, we configure the bounds as follows:
English ($\mathcal{B}_{En}\in[0.8, 0.9]$) is set below parity to force active summarization;
German ($\mathcal{B}_{De}\in[0.9, 1.0]$) maintains relative pressure despite lower density;
and Spanish ($\mathcal{B}_{Es}\in[1.0, 1.1]$) significantly reduces the unconstrained baseline of $1.84\times$ to ensure intelligibility under strict time limits while respecting its information density.

\begin{table}[t]
\centering
\small
\setlength{\tabcolsep}{2.5pt}
\begin{tabular}{lccc|c}
\toprule
\textbf{Target} & \textbf{Theoretical} & \textbf{LLM Baseline} & \textbf{Bias} & \textbf{Ours (Target)} \\
\textbf{Lang} & ($\mu_{theo}$) & ($\mu_{LLM}$) & ($\Delta$) & ($\mathcal{B}_L$) \\
\midrule
En & $1.03\times$ & $1.35\times$ & $+0.32$ & $\mathbf{0.8} \sim \mathbf{0.9}$ \\
De & $1.19\times$ & $1.59\times$ & $+0.40$ & $\mathbf{0.9} \sim \mathbf{1.0}$ \\
Es & $1.49\times$ & $1.84\times$ & $+0.35$ & $\mathbf{1.0} \sim \mathbf{1.1}$ \\
\bottomrule
\end{tabular}
\vspace{-2mm}
\caption{Comparison of expansion ratios. Target Bounds ($\mathcal{B}_L$) are set to remove model bias ($\Delta$) and enforce semantic condensation.}
\label{tab:expansion_comparison}
\vspace{-3mm}
\end{table}

%% file: section/Method.tex
\section{Method}

\begin{figure*}[t]
\centering
\includegraphics[width=0.85\textwidth]{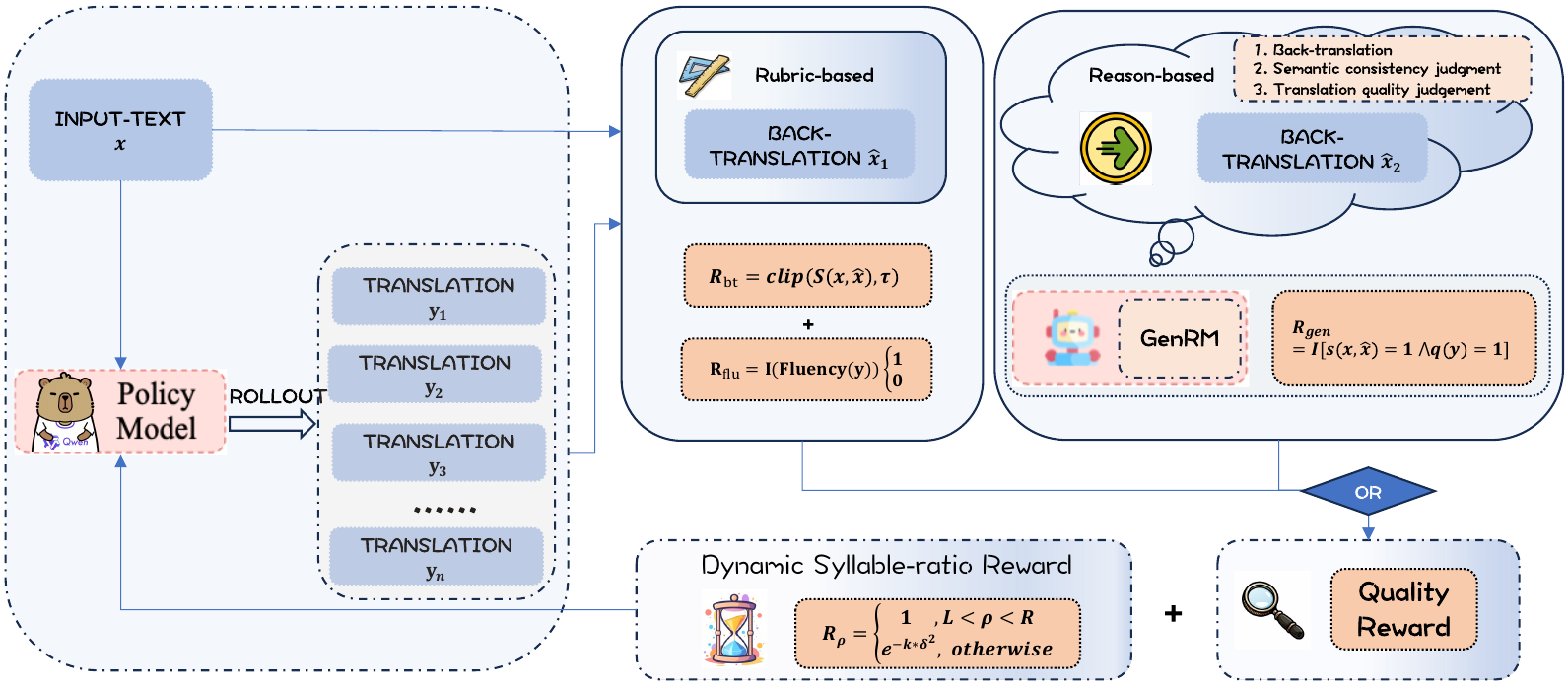}
\caption{Overview of our \textsc{Homura} framework. Given a source utterance $x$ and a syllable-ratio budget $c$, we optimize the policy $\pi_\theta(y\!\mid\!x,c)$ with GRPO. The reward combines a length-compliance term $R_{\mathrm{len}}$ and a quality term $R_{\mathrm{qual}}$ (rubric-based or reason-based), encouraging concise yet faithful translations.}
\label{fig:method}
\vspace{-2mm}
\end{figure*}

We formulate syllable-level control as a standard controllable text generation task\cite{li2024reinforcement, gu2025length}, where our objective is to maximize translation quality while satisfying the prescribed syllable-ratio constraint. As illustrated in Figure~\ref{fig:method}, we optimize the model with GRPO~\citep{shao2024deepseekmath}, a lightweight PPO-style algorithm designed for sequence-level rewards.

\subsection{Reward Design}
\label{sec:reward}
In our setting, the control condition $c$ corresponds to satisfying a strict length budget while maintaining semantic consistency with the source. This requires a reward that simultaneously encourages length compliance and discourages meaning drift. 
We instantiate this task reward as:
\begin{align}
R(x,c,y)
= \lambda_{\mathrm{len}} \, R_{\mathrm{len}}(x,y)
+ \lambda_{\mathrm{qual}} \, R_{\mathrm{qual}}(x,y),
\end{align}
where $x$ is the source utterance, $y$ is the generated translation, and $\lambda_{\mathrm{len}}, \lambda_{\mathrm{qual}} \ge 0$ control the trade-off between temporal alignment and semantic fidelity.

\subsubsection{Dynamic Syllable-ratio Reward}
\label{sec:dynamic_reward}

To ensure the translation length aligns naturally with the source speech across varying scales, we propose a dynamic syllable-ratio reward. The syllable ratio $\rho$ for a source-target pair $(x, y)$ is defined as:
\begin{equation}
\rho(x,y) = \frac{\sigma(y)}{\max(\sigma(x), 1)},
\end{equation}
where $\sigma(\cdot)$ is the syllable count of a sequence.

Recognizing that strict length constraints must be sensitive to both linguistic density and utterance length, we implement a dynamic acceptance interval $[L(x), R(x)]$. We initialize the baseline bounds $[L_0, R_0]$ using the language-specific Target Bounds $\mathcal{B}_L$ set in Section~\ref{sec:verbosity_bias}. This ensures the optimization target respects the cross-lingual density hierarchy established in our theoretical analysis.

However, for shorter utterances, the ``granularity'' of syllable counting introduces high variance (discretization noise), making rigid bounds brittle. To address this, for utterances where the source length $\sigma(x)$ is below the corpus mean $\mu$, we define a scaling factor $\gamma(x)$ to adaptively relax the lower threshold:
\begin{equation}
\gamma(x) = \alpha_1 + \alpha_2 \left( \frac{\sigma(x)}{\mu} \right)^{1/2},
\end{equation}
where $\alpha_1, \alpha_2$ are hyperparameters governing the degree of relaxation. The final dynamic bounds are formulated as $[L(x), R(x)] = [L_0 \cdot \gamma(x), \, R_0]$.

To facilitate stable policy optimization, we avoid sparse or binary signals in favor of a smoother reward landscape. The reward function $R_{\mathrm{len}}(x,y)$ evaluates compliance with these dynamic boundaries via a squared exponential decay:
\begin{equation}
R_{\mathrm{\rho}} = 
\begin{cases} 
1 & \text{if } \rho \in [L(x), R(x)] \\
\exp(-k \delta^2) & \text{otherwise}
\end{cases},
\end{equation}
where $\delta(x,y) = \max\bigl(|\rho(x,y) - L(x)|, |\rho(x,y) - R(x)|\bigr)$ denotes the deviation. This non-linear formulation provides a soft penalty that effectively guides the model toward the density-calibrated region while preventing training instability associated with hard-truncation rewards. In Appendix~\ref{sec:appendix_dynamic_reward}, we further investigate the effectiveness of a static syllable-ratio reward.

\subsubsection{Rubric-based Quality Reward}
We adopt a two-part quality rubric\cite{hashemi2024llm, liu2023g}, covering (i) semantic fidelity and (ii) linguistic well-formedness, and implement it as the following reward components.

\paragraph{Semantic Fidelity Reward}
\label{sec:sem_reward}

Inspired by the probabilistic duality in \citep{he2016dual}, we enforce content preservation via a reconstruction-based reward. We employ a frozen \texttt{DeepSeek-V3} \citep{deepseekai2024deepseekv3technicalreport} to back-translate the hypothesis $y$ into $\hat{x} = f_{bt}(y)$. To measure semantic retention, we compute the cosine similarity $S(x, \hat{x})$ between embeddings of the source $x$ and reconstruction $\hat{x}$ extracted by \texttt{Qwen-Embedding-0.6B} \citep{qwen3embedding}:

\begin{equation}
    S(x, \hat{x}) = \frac{\mathbf{v}_x \cdot \mathbf{v}_{\hat{x}}}{\|\mathbf{v}_x\| \|\mathbf{v}_{\hat{x}}\|}.
\end{equation}

The final reward is clipped to stabilize optimization:
\begin{equation}
    R_{bt}(x, y) = \text{clip}(S(x, \hat{x}), \tau_{min}, \tau_{max}).
\end{equation}

\paragraph{Linguistic Well-formedness Reward}
\label{sec:ling_reward}

To prevent the telegraphic outputs often associated with aggressive compression, we explicitly penalize linguistic degradation. We utilize a frozen \texttt{DeepSeek-V3} as a judge to evaluate translations against a rubric covering: (i) \textit{grammaticality}, (ii) \textit{readability}, and (iii) \textit{language consistency} (e.g., no code-switching). The model outputs a binary decision $s_{flu} \in \{0, 1\}$, serving directly as the reward:

\begin{equation}
    R_{flu}(x, y) = s_{flu}.
\end{equation}

Full prompting details are in Appendix~\ref{sec:prompt_fluency}.


\subsubsection{Reason-based Quality Reward}
Beyond the simple compositional rewards, we explore an end-to-end generative reward model (GenRM)\cite{mahan2024generative, liang2025generative} that evaluates translation quality via explicit reasoning.
GenRM is trained using structured Chain-of-Thought (CoT) annotations synthesized by \texttt{Gemini-2.5-Pro} \citep{comanici2025gemini}. To ensure high-quality reasoning, we filter the raw CoT data using the aforementioned rubric-based method. The model is instantiated from \texttt{Qwen3-32B}~\citep{yang2025qwen3} and optimized via supervised fine-tuning (SFT) over the full CoT sequence and the final decision token.

Given a source sentence $x$ and a translation hypothesis $y$, GenRM performs a single long CoT reasoning process that can be abstracted as
\begin{equation}
\hat{x} = f_{\mathrm{bt}}(y),\;
s(x, \hat{x}), q(y) \in \{0,1\},
\end{equation}
where $f_{\mathrm{bt}}$ denotes implicit back-translation, $s(x, \hat{x})$ indicates whether the core information in $x$ is preserved, and $q(y)$ evaluates the intrinsic quality of $y$ in terms of adequacy and fluency. Based on these decisions, GenRM outputs a binary quality reward
\begin{equation}
R_{gen}(x,y) = \mathbb{I}\big[s(x,\hat{x}) =1 \land q(y)=1\big].
\end{equation}
This formulation yields a reference-free, interpretable quality signal and enforces an explicit conjunction between information preservation and translation quality. Concrete examples of the CoT reasoning are provided in the Appendix \ref{sec_genrm_example_appendix}.
Ablation studies on the reward components and design choices are provided in Appendix~\ref{sec:appendix_reward_ablation}.

%% file: section/Experiment.tex
\section{Experiments and Results}
\label{sec_experiment}

\begin{table*}[t]
    \centering
    \caption{Performance Comparison on Sand-Glass Across All Language Pairs ($\text{BLEU-$\rho$}$, $\text{Cometkiwi}$, $\text{BT-CERR}$).  All results are averaged by 3 times.}
    \label{tab:main_results_cross_lingual}

    \setlength{\tabcolsep}{3pt}
    \renewcommand{\arraystretch}{1.1}

    \resizebox{\textwidth}{!}{%
    \begin{tabular}{l|cccccc|cccccc|cccccc}
        \toprule
        \multirow{3}{*}{\textbf{Model}} & \multicolumn{18}{c}{\textbf{Evaluation Metrics on Sand-Glass}} \\
        \cmidrule(lr){2-19}
         & \multicolumn{6}{c|}{\textbf{Zh $\rightarrow$ En}} & \multicolumn{6}{c|}{\textbf{Zh $\rightarrow$ De}} & \multicolumn{6}{c}{\textbf{Zh $\rightarrow$ Es}} \\
        \cmidrule(lr){2-7} \cmidrule(lr){8-13} \cmidrule(lr){14-19}
          &  \textbf{IB}  &  $\rho$  &  \textbf{CK}  &  \textbf{BC}  &  \textbf{B-$\rho$}  &  \textbf{Tok.}  &  \textbf{IB}  &  $\rho$  &  \textbf{CK}  &  \textbf{BC}  &  \textbf{B-$\rho$}  &  \textbf{Tok.}  &  \textbf{IB}  &  $\rho$  &  \textbf{CK}  &  \textbf{BC}  &  \textbf{B-$\rho$}  &  \textbf{Tok.} \\
        \midrule
        \multicolumn{19}{l}{\textit{Category 1: Unconstrained LLMs}} \\
        \midrule
        \texttt{claude-4.1-opus}  &  22.4\%  &  1.361  &  0.741  &  0.943 &  0.313 &  17.1 &  12.1\%  &  1.582  &  0.676&  0.882 &  0.247 &  24.2 &  8.2\%  &  1.812  &  0.668&  0.900 &  0.220 &  21.3 \\
        \texttt{deepseek-v3}  &  22.6\%  &  1.413  &  0.748  &  0.941 &  0.287 &  46.3 &  12.4\%  &  1.674  &  0.676&  0.889 &  0.222 &  62.3 &  9.1\%  &  1.990  &  0.671&  0.884 &  0.191 &  99.6 \\
        \texttt{gemini-2.5-pro}$^*$  &  24.9\%  &  1.289  &  0.739  &  0.941 &  0.307 &  12.1 (916.9) &  15.5\%  &  1.511  &  0.666&  0.877 &  0.238 &  14.2 (125.8) &  9.9\%  &  1.718  &  0.668&  0.883 &  0.212 &  14.0 (1532.7) \\
        \texttt{gpt-5}$^*$  &  23.2\%  &  1.330  &  0.746  &  0.947 &  0.300 &  22.4 (412.8) &  11.0\%  &  1.633  &  0.681&  0.894 &  0.220 &  23.2 (605.6) &  8.3\%  &  1.862  &  0.671&  0.899 &  0.206 &  24.0 (676.8) \\

        \midrule
        \multicolumn{19}{l}{\textit{Category 2: Prompt-based Compression}} \\
        \midrule
        \texttt{claude-4.1-opus}  &  46.1\%  &  0.852  &  0.691&  0.907 &  0.376&  12.2 &  31.5\%  &  0.951  &  0.535 &  0.838 &  0.293 &  17.2 &  26.6\%  &  1.106  &  0.631&  0.849 &  0.261 &  13.5 \\
        \texttt{deepseek-v3}  &  38.4\%  &  0.960  &  0.700&  0.924 &  0.307 &  12.4 &  20.3\%  &  1.196  &  0.557 &  0.866 &  0.245 &  15.8 &  16.1\%  &  1.355  &  0.556 &  0.868 &  0.225 &  16.3 \\
        \texttt{gemini-2.5-pro}$^*$  &  49.7\%  &  0.920  &  0.698&  0.927 &  0.322 &  10.7 (1231.3) &  52.3\%  &  1.016  &  0.517 &  0.853 &  0.266 &  12.3 (1450.1) &  46.7\%  &  0.974  &  0.586&  0.827 &  0.252 &  10.7 (554.5) \\
        \texttt{gpt-5}$^*$  &  53.6\%  &  0.937  &  0.705&  0.928 &  0.345 &  21.7 (2024.9) &  51.5\%  &  1.025  &  0.534 &  0.874 &  0.283 &  23.1 (1962.3) &  62.6\%  &  0.982  &  0.586&  0.833 &  0.270 &  21.0 (3105.8) \\

        \midrule
        \multicolumn{19}{l}{\textit{Category 3: Best-of-N Refinement}} \\
        \midrule
        \texttt{claude-4.1-opus}  &  67.6\%  &  0.845  &  0.691&  0.903 &  0.367 &  109.7 &  50.9\%  &  0.946  &  0.611 &  0.848 &  0.324 &  109.2 &  45.5\%  &  1.039  &  0.621&  0.857 &  0.304 &  110.0 \\
        \texttt{deepseek-v3}  &  63.6\%  &  0.852  &  0.691&  0.902 &  0.332 &  122.6 &  44.7\%  &  0.959  &  0.527 &  0.863 &  0.286 &  160.5 &  44.4\%  &  1.042  &  0.522&   0.851 &  0.275 &  113.6 \\
        \texttt{gemini-2.5-pro}$^*$  &  61.9\%  &  0.900  &  0.673&  0.904 &  0.276 &  133.2 (1089.2) &  44.1\%  &  0.992  &  0.517 &  0.854 &  0.241 &  136.0 (972.7) &  42.3\%  &  1.087  &  0.554&  0.858 &  0.235 &  133.8 (611.1) \\
        \texttt{gpt-5}$^*$  &  67.2\%  &  0.888  &  0.673&  0.906 &  0.296 &  114.2 (1810.2) &  47.4\%  &  0.994  &  0.571 &  0.861 &  0.275 &  111.2 (2229.6) &  51.5\%  &  1.076  &  0.529&  0.855 &  0.256 &  110.8 (2158.8) \\

        \midrule
        \multicolumn{19}{l}{\textit{Category 4: Pipeline-based Post-editing (2 run)}} \\
        \midrule
        \texttt{claude-4.1-opus}  &  38.3\%  &  0.918  &  0.694&  0.904 &  0.327 &  30.2 &  24.1\%  &  1.089  &  0.653&  0.868 &  0.276 &  42.7 &  18.7\%  &  1.284  &  0.643&  0.876 &  0.246 &  37.6 \\
        \texttt{deepseek-v3}  &  38.4\%  &  0.967  &  0.703&  0.934 &  0.322 &  88.1 &  23.3\%  &  1.154  &  0.645&  0.874 &  0.239 &  89.4 &  17.8\%  &  1.301  &  0.543& 0.873 &  0.221 &  102.1 \\
        \texttt{gemini-2.5-pro}$^*$  &  50.0\%  &  0.837  &  0.691&  0.902 &  0.348 &  22.3 (2517.7) &  41.9\%  &  1.001  &  0.627&  0.830 &  0.267 &  27.5 (1043.1) &  35.2\%  &  1.042  &  0.602&  0.840 &  0.250 &  22.4 (2975.2) \\
        \texttt{gpt-5}$^*$  &  78.6\%  &  0.833  &  0.690&  0.917 &  0.321 &  42.2 (2666.2) &  75.5\%  &  0.934  &  0.514&  0.842 &  0.275 &  46.6 (2773.0) &  66.3\%  &  1.018  &  0.557&  0.820 &  0.234 &  44.9 (4012.9) \\
        \midrule
        \multicolumn{19}{l}{\textit{Category 5: \textsc{Homura}}} \\
        \midrule
        \textbf{Ours (\textsc{Homura}\textsubscript{Rubric})}  &  86.3\%  &  0.809  &  0.700 &  0.914 &  0.378 &  9.2  &  84.3\%  &  0.901  & 0.591 &   0.853 &  0.317 &  16.3&   89.0\%  &  1.067  &   0.620& 0.863 &  0.260 &  13.1\\
        \textbf{Ours (\textsc{Homura}\textsubscript{Reason})}  &  89.6\%  &  0.846  &  0.701 &  0.925 &  0.376 & 10.1&  84.6\%  &  0.918  & 0.603 &   0.861 &  0.321  & 14.7&  87.3\% &  1.057  &  0.605 &  0.858 &  0.309 &  13.2 \\
        \bottomrule
    \end{tabular}%
    }
    \vspace{2mm}

    \raggedright
    \footnotesize
    Note: \textbf{IB}: In-Bounds Rate (\%). $\rho$: syllable ratio. \textbf{CK}: CometKiwi. \textbf{BC}: BT-CERR. \textbf{B-$\rho$}: BLEU-$\rho$. \textbf{Tok.}: avg output tokens; for reasoning models ($^*$), parenthesized values include internal thinking tokens.

    \vspace{-4mm}
\end{table*}

\subsection{Experimental Setup}

We evaluate SOTA LLMs under five strategy categories that span the length-control spectrum:

\begin{itemize}[nosep]
    \item \textbf{Category 1: Unconstrained LLMs}. Frontier models (\texttt{GPT-5}~\cite{openai2025gpt5systemcard}, \texttt{Claude-4.1-Opus}~\cite{anthropic2025claude}, \texttt{Gemini-2.5-Pro}, \texttt{DeepSeek-V3}) with standard decoding. For \texttt{GPT-5} and \texttt{Gemini-2.5-Pro}, we set low reasoning effort to approximate direct translation. These serve as semantic upper bounds.
    \item \textbf{Category 2: Prompt-based Compression.} Category 2 models with brevity-oriented system prompts enforcing syllable-level conciseness.
    \item \textbf{Category 3: Best of N Refinement.} Generating $N$ candidates with varying lengths, followed by a greedy selection of the highest-fidelity output that fulfills the syllable budget.
    \item \textbf{Category 4: Pipeline-based Post-editing.} Two-stage translation $\rightarrow$ length-constrained rewriting.
    \item \textbf{Category 5: Our Method (\textsc{Homura}).} RL fine-tuning on \texttt{Qwen3-8B}~\citep{yang2025qwen3} to enforce temporal budgets as hard constraints.
\end{itemize}

All baseline prompt templates are provided verbatim in Appendix~\ref{app:baseline_prompts}. More experimental details are provided in Appendix~\ref{sec:appendix_hyper}. 

\subsection{Target Syllable Bounds Configuration}
\label{ssec:target_bounds}
Following Section~\ref{sec:verbosity_bias}, we set density-calibrated target bounds $\mathcal{B}_{L}$ to impose comparable compression pressure across languages: En $\in [0.8, 0.9]$, De $\in [0.9, 1.0]$, and Es $\in [1.0, 1.1]$. These intervals are used as hard constraints for \textsc{Homura}.

\subsubsection{Evaluation Metrics}
\label{sec:evaluation_metrics}
We report complementary metrics for efficiency, semantic preservation, and linguistic quality:
\paragraph{IB: In-Bounds Rate (\% $\uparrow$)}
We define the \textbf{In-Bounds Rate} (\textbf{IB}) as the percentage of translations that satisfy the target interval bounds:
\begin{equation}
    \text{IB} = \frac{1}{|D|} \sum_{(x, \mathcal{B}_L) \in D} \mathbb{I}\left( \frac{\sigma(y)}{\sigma(x)} \in \mathcal{B}_L \right),
\end{equation}
where $D$ is the evaluation dataset, and $\mathbb{I}(\cdot)$ is the indicator function.
\paragraph{Translation Density (BLEU-$\rho$ $\uparrow$)}
We define \textbf{BLEU-$\rho$} as semantic fidelity per temporal unit:
\begin{equation}
    \text{BLEU-}\rho = \frac{\text{BLEU}(x, \hat{x})}{\rho(x,y)},
\end{equation}
where $\text{BLEU}(x,\hat{x})$ is the back-translation BLEU~\cite{papineni2002bleu}. Since BLEU alone may not fully capture fluency in back-translation paradigms~\cite{edunov-etal-2020-evaluation}, we use BLEU-$\rho$ specifically to measure \textbf{information density}, while linguistic quality assessed via Cometkiwi.

\paragraph{Semantic Integrity (BT-CERR $\uparrow$)}
BT-CERR is a binary indicator that evaluates whether all core events extracted from $x$ are retained in the back-translation $\hat{x}$.

\paragraph{Linguistic Quality (Cometkiwi $\uparrow$)}
We use reference-free Cometkiwi\cite{rei2022cometkiwi} to assess fluency and adequacy under compression.

\subsection{Main Results}
\label{sec:main_results}

\subsubsection{Cross-Lingual Performance and Temporal Compliance}
\label{sec:overall_performance}

Table~\ref{tab:main_results_cross_lingual} summarizes the main results on \textsc{Sand-Glass}, with additional zero-shot generalization results on WMT24pp\cite{deutsch2025wmt24++} and \textsc{MuST-Cinema}\cite{karakanta2020must} reported in Appendix~\ref{sec:generalization_wmt}. We report human evaluation in Appendix~\ref{sec:appendix_human_judgment} and Statistical Significance Testing in Appendix~\ref{sec_statistical_test_appendix}. Overall, the results highlight the trade-off between temporal compliance and semantic density.

\paragraph{Budget Violation in Frontier Models.} 
As shown by the $\rho$ column, unconstrained frontier models(\textsc{Cat 1}) universally fail to meet temporal budgets, particularly in low-density languages like Spanish. While specialized prompting (\textsc{Cat 2}) shifts the distribution toward brevity, it remains unreliable, with models like \texttt{GPT-5} still exhibiting "verbosity drift" that exceeds target bounds. This confirms that pure instruction-following is insufficient for rigorous syllable-level constraints.

\paragraph{The Search-Fidelity Trade-off.} 
Search-based (\textsc{Cat 3}) and pipeline (\textsc{Cat 4}) strategies achieve higher In-Bounds rates (IB) but at a significant semantic cost. For instance, \textsc{Cat 4} variants often show a drop in BT-CERR, suggesting that the disjointed post-editing process leads to "hallucinated omissions" of critical source predicates. Moreover, the increased inference latency of Best-of-$N$ makes it less practical in real-time settings.

\begin{figure*}[th]
    \centering
    \begin{subfigure}[t]{0.45\linewidth}
        \centering
        \includegraphics[width=\linewidth]{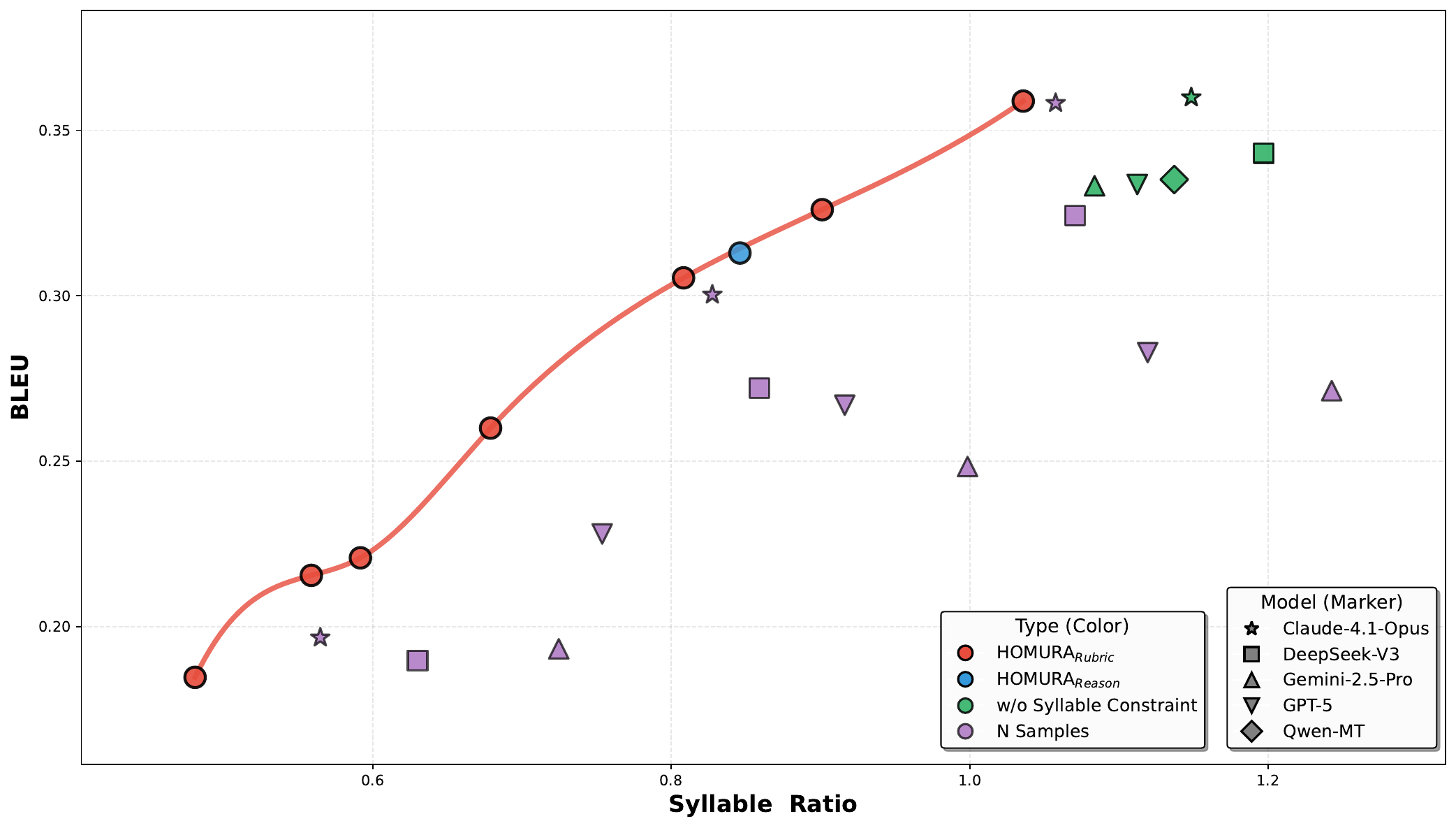}
        \caption{BLEU vs. Avg. Syllable Ratio}
        \label{fig:bt_bleu}
    \end{subfigure}\hfill
    \begin{subfigure}[t]{0.45\linewidth}
        \centering
        \includegraphics[width=\linewidth]{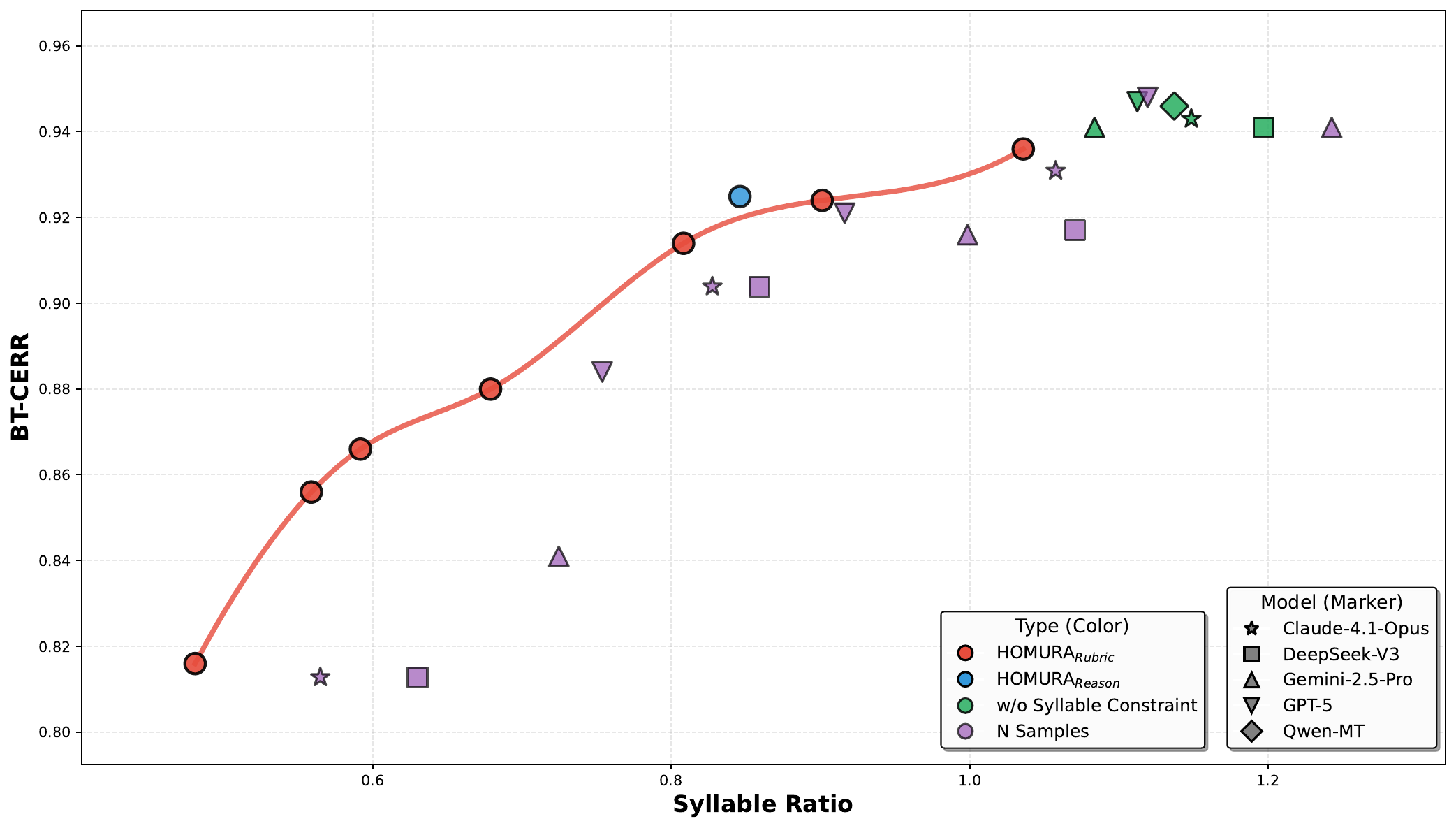}
        \caption{BT-CERR vs. Avg. Syllable Ratio}
        \label{fig:cerr_en}
    \end{subfigure}
    \caption{Quality--compression trade-off on \text{Zh}$\rightarrow$\text{En}. We plot BLEU (left) and BT-CERR (right) versus the average syllable ratio $\rho$. Each point is a model setting (marker: backbone; color: variant: Unconstrained, Rubric, GenRM, Multi-length Prompting). The red curve shows the empirical trend: quality degrades as compression strengthens.}
    
    \label{fig:tradeoff_curve}

    \vspace{-4mm}
\end{figure*}

\paragraph{Information Density and Superior Compliance of \textsc{Homura}.}
Concurrently, \textsc{Homura} achieves up to an 89.6\% temporal compliance rate while achieving the highest BLEU-$\rho$ across all language pairs, demonstrating active \textit{semantic packing} that maximizes the semantic payload per syllable. We further validate the practical deliverability and scalability of our framework via two key analyses detailed in the Appendix: (1) an evaluation of Syllables Per Second (SPS) and speech-rate thresholds, which confirms that \textsc{Homura} achieves the lowest average SPS and highest pass rate for comfortable video articulation (Appendix~\ref{sec:sps}); and (2) a scaling analysis demonstrating consistent gains with a 32B model (Appendix~\ref{sec:appendix_scale}).

\paragraph{Inference efficiency.}
\textsc{Homura} is markedly more inference-efficient. Unlike Best-of-$N$, which multiplies decoding cost by generating $N$ candidates, and unlike translate-then-modify pipelines that require an extra rewriting pass, \textsc{Homura} satisfies the syllable/temporal budget in a single decoding. Consequently, it attains high temporal compliance with substantially fewer tokens (AVG-Tokens in Table~\ref{tab:main_results_cross_lingual}), making it suitable for real-time and high-throughput deployment.

\paragraph{Robustness of Reason-based GenRM.} 
Within Category~5, \textsc{Homura}\textsubscript{Reason} achieves higher BT-CERR than \textsc{Homura}\textsubscript{Rubric} on two of the three target languages, with the largest gain observed for English.
This suggests the reason-based reward better prunes redundancy while preserving the source’s logical structure. 
We further validate this by replacing the self-trained GenRM in \textsc{Homura}\textsubscript{Reason} with an LLM-as-RM baseline (\texttt{DeepSeek-V3}); results are in Appendix~\ref{sec:appendix_genrm_ablation}.





\subsubsection{Analysis of Quality--Compression Trade-off}
\label{sec:compression_tradeoff}

Figure~\ref{fig:tradeoff_curve} illustrates the trade-off between translation quality and output length by plotting semantic metrics against the average syllable ratio $\rho$.

As shown, unconstrained baselines cluster in the high-quality, high-$\rho$ region. Although standard LLMs can be prompted to shorten outputs, their quality drops sharply under aggressive compression. This reflects a low rate--distortion efficiency: when forced to reduce the phonetic ``rate'' (syllables), these models suffer from severe semantic ``distortion'' due to suboptimal content selection.

In contrast, \textsc{Homura} demonstrates superior rate--distortion efficiency, maintaining higher semantic fidelity at any given syllable budget. Rather than executing naive truncation, \textsc{Homura} actively optimizes the generation trajectory to operate at a more efficient trade-off point, packing a higher semantic payload into the limited phonetic constraints required by media localization.

\subsubsection{Qualitative Case Studies}
\label{sec:case_studies}

Table~\ref{tab:case_study} qualitatively benchmarks \textsc{Homura} against \texttt{Claude-4.1-Opus} baselines. Unlike baseline strategies that suffer from over-translation, syntactic fragmentation, or a coarse \emph{cliff pattern}, \textsc{Homura} reliably hits the target window in a single pass while preserving pragmatic integrity (full analyses are deferred to Appendix~\ref{sec:appendix_cases}).

\subsection{Autonomous Compression: Navigating the Compression Limit}
\label{sec:auto_compression}

The preceding experiments show that \textsc{Homura} can reliably translate under a user-specified ratio interval. We further ask whether, without a preset lower bound, the model can autonomously discover the practical compression limit: the point where additional shortening triggers a sharp loss in meaning. To this end, we formulate an autonomous compression task by replacing the hard-interval length reward with a continuous incentive:
\begin{align}
R_{\mathrm{len}}^{\mathrm{(auto)}}(x, y) = \min\left(\cos\left(\theta \cdot \rho(x, y)\right),\; 0\right),
\end{align}
where $\theta$ scales the incentive gradient. The policy is encouraged to minimize $\rho$ while still being constrained by the quality reward $R_{\mathrm{qual}}$.

\begin{table}[t] 
\centering 
\caption{The Compression Wall. The autonomous model reliably reaches the feasible limit; stricter manual targets miss $\rho$ and produce worse compression.}
\label{tab:auto_compression_results} 
\small
\setlength{\tabcolsep}{10pt}
\begin{tabular}{lccc} \toprule
\textbf{Model \& Configuration} & $\boldsymbol{\rho}$ & \textbf{BLEU-$\boldsymbol{\rho}$} & \textbf{BT-CERR}\\
\midrule \textbf{Auto \textsc{Homura}} & \textbf{0.485} & 0.404 & 0.811 \\ 
\midrule \multicolumn{4}{l}{\textit{Manually Constrained \textsc{Homura}}} \\ 
\hspace{1mm} $R_{\mathrm{target}} \in [0.3, 0.4]$ & 0.558 & 0.392 & 0.819\\
\hspace{1mm} $R_{\mathrm{target}} \in [0.4, 0.5]$ & \textbf{0.481} & 0.384 & 0.816\\ 
\hspace{1mm} $R_{\mathrm{target}} \in [0.5, 0.6]$ & 0.559 & 0.386 & 0.856\\
\hspace{1mm} $R_{\mathrm{target}} \in [0.8, 0.9]$ & 0.808 & 0.378 & 0.914\\ 
\bottomrule
\end{tabular}
\vspace{-4mm}
\end{table}




\paragraph{The Empirical Compression Wall.}
We observe a stable \textit{compression wall} when pushing the model to its structural limit. As detailed in Table~\ref{tab:auto_compression_results}, autonomous optimization consistently converges to a narrow operating regime around $\rho \in [0.46, 0.49]$, regardless of the reward configuration. Crucially, when we enforce an ultra-aggressive manual target of $[0.3, 0.4]$, the model does \emph{not} become shorter; instead, it rebounds to $\rho=0.558$ while semantic coherence collapses. This non-linear failure indicates a rigid \textit{feasibility threshold} around $\rho \approx 0.49$ for Zh$\rightarrow$En translation, beyond which the model cannot maintain basic syntactic well-formedness.

\paragraph{Rate-Distortion Efficiency and Feasibility Boundary.}
To quantify this behavior, Auto \textsc{Homura} attains the optimal $\text{BLEU-}\rho$ ($0.404$) while operating at a genuinely shorter length. Qualitative analysis reveals that autonomous optimization navigates this quality--compression frontier via active \emph{semantic condensation} (e.g., stripping non-essential modifiers and deploying dense lexical substitutions). Conversely, forcing the model into an unnatural manual interval triggers unstable truncation, reflecting the severe tension between mechanical length reduction and core meaning preservation. This empirical wall effectively represents the \textit{minimal information carrier} required to express core predicate--argument content, closely aligning with the theoretical density floor predicted by the Iso-Information Principle (Section~\ref{sec:verbosity_bias}).

%% file: section/Conclusion.tex
\section{Conclusion}
\label{sec:conclusion}

We address systemic cross-lingual verbosity bias in LLM translation, a key obstacle for time-constrained scenarios such as subtitling and dubbing. To study this problem, we introduce \textsc{Sand-Glass}, a syllable-budgeted benchmark grounded in the Iso-Information Principle, and propose \textsc{Homura}, a constrained RL framework that optimizes the trade-off between semantic fidelity and temporal feasibility.

\textsc{Homura} uses a dynamic syllable-ratio reward to encourage translations that remain within the prescribed budget while preserving meaning, without relying on reference translations. Experiments show that \textsc{Homura} substantially improves length compliance and information density, as measured by BLEU-$\rho$. Future work will examine whether the observed compression feasibility boundary, e.g., $\rho \approx 0.49$ for Zh$\rightarrow$En, generalizes across language pairs, and extend \textsc{Homura} to end-to-end speech translation with joint semantic and duration optimization.

%% file: section/Appendix.tex
\section{Related Work}
\label{sec:appendix_related_works}
\subsection{Constrained Translation and Subtitling}
Media localization imposes strict spatiotemporal constraints beyond semantic fidelity \citep{karakanta-etal-2020-must}. To address these limitations, early research focused on supervised learning paradigms that integrate specialized markers to guide line length and reading speed \citep{karakanta-etal-2020-42}. An advancement was introduced by \citet{wu-etal-2023-videodubber} with \textit{VideoDubber}, which pioneered direct speech-length control by incorporating token-level duration predictions into the decoder stack. However, this approach remains constrained by its dependence on supervised training with non-isochronous datasets and heuristic duration labeling that introduces annotation noise. Building on this, \citet{chadha2025lengthawarespeechtranslation} demonstrated that even with automatically generated length tags derived from phoneme ratios, supervised training on non-isochronous speech-translation data remains a fundamental bottleneck for achieving precise isochronicity. While architectures like CLSC \citep{jorgensen-mengshoel-2025-cross} further refine this task using control tokens to manage sequence length, they remain fundamentally dependent on supervised training with task-specific compression corpora. In contrast, our work targets general-purpose Large Language Models (LLMs) via a reinforcement learning framework. This approach explicitly optimizes the trade-off between semantic preservation and temporal compliance without requiring ground-truth compression datasets. Recently, the focus has shifted toward leveraging general-purpose Large Language Models (LLMs). For instance, \citet{javorsky-etal-2025-prompting} investigated prompt-based strategies for machine translation under strict isometric constraints, highlighting the potential of in-context learning for length control. In contrast, our work moves beyond pure prompting paradigms by introducing a reinforcement learning framework. This approach explicitly optimizes the trade-off between semantic preservation and temporal compliance without requiring ground-truth compression datasets or fragile prompt engineering.

\subsection{Hard Constraints and Lyric Translation}
The imposition of strict feasibility constraints parallels automatic lyric translation, where models must adhere to syllable counts and rhythm \citep{ou-etal-2023-songs}. Building on curriculum RL strategies for discrete constraints \citep{ren2025lyricar}, \textsc{Homura} treats temporal duration as a hard constraint. We balance these rigid length requirements with semantic quality, ensuring the model respects the "sand-glass" temporal budget.

\subsection{Verbosity Bias in LLMs}
A major challenge for time-constrained translation is the systemic "verbosity bias" in LLMs, often exacerbated by RLHF alignment that conflates length with helpfulness \citep{saito2023verbosity, skalse2022defining, singhal2023long, zhao2026saber}. Our framework counteracts this by introducing a dynamic syllable-ratio reward. Instead of encouraging length drift, we specifically penalize verbosity, aligning the model's output with the strict density requirements of subtitling and dubbing.

\section{Statistical Significance Testing}
\label{sec_statistical_test_appendix}

We perform hypothesis testing to examine whether the improvements brought by \textsc{Homura} are statistically significant.
For a fair comparison, for each baseline we choose its best-performing single-pass configuration and compare it with \textsc{Homura} under the same setting.
Table~\ref{tab:statistical_sign_test} reports the resulting p-values.

\begin{table}[th]
    \centering
    \caption{p-values for testing \textsc{Homura} against each variant.}
    \label{tab:statistical_sign_test}
    
    \setlength{\tabcolsep}{8pt}
    \renewcommand{\arraystretch}{1.1}
    \small

    \begin{tabular}{l|cc|cc}
        \toprule
        \textbf{Model} & \multicolumn{4}{c}{\textbf{p-value of \textsc{Homura}}} \\
        \cmidrule(lr){2-5}
        & \multicolumn{2}{c|}{\textbf{Zh $\rightarrow$ En}} & \multicolumn{2}{c}{\textbf{Zh $\rightarrow$ De}} \\
        \cmidrule(lr){2-3}\cmidrule(lr){4-5}
        & \textbf{B-$\rho$} & \textbf{CK}
        & \textbf{B-$\rho$} & \textbf{CK} \\
        \midrule
        \textbf{Homura\textsubscript{Rubric}}
          & 6.46e-03 & 8.20e-03
          & $<\!1e\!-\!10$ & $<\!1e\!-\!10$ \\
        \textbf{Homura\textsubscript{Reason}}
          & - & 6.30e-03
          & $<\!1e\!-\!10$ & $<\!1e\!-\!10$ \\
        \bottomrule
    \end{tabular}
    \vspace{2mm}
\end{table}

Across both language pairs and evaluation metrics, \textsc{Homura} yields statistically significant improvements over its strongest single-pass baselines, with p-values consistently below the conventional 0.01 threshold (often far smaller, e.g., $<\!1e\!-\!10$ for Zh$\rightarrow$De). These results confirm that the observed gains are unlikely to be due to random variation. The ``--'' entry denotes a degenerate case where the two compared results are identical, yielding no measurable difference and thus no meaningful significance test.

\section{Human Evaluation}
\label{sec:appendix_human_judgment}
We additionally report human evaluation results for the Zh-to-En setting to further validate the effectiveness of our method and the appropriateness of our evaluation metrics.

\begin{table}[h]
\centering
\caption{Human evaluation results on Accuracy and Completeness (Zh $\rightarrow$ En).}
\label{tab:human_eval_acc_comp}
\setlength{\tabcolsep}{6pt}
\renewcommand{\arraystretch}{1.1}
\small
\begin{tabular}{@{}p{2.6cm}lcccccc@{}}
\toprule
\textbf{Setting} & \textbf{Model} & \textbf{IB} & $\rho$ & \textbf{Acc.} & \textbf{Comp.} & \textbf{Avg.}\\
\midrule
\multirow{4}{*}{\shortstack[l]{Unconstrained\\LLMs}}
& Gemini-2.5-Pro   & 24.9\% & 1.289 & 4.95 & 5.00 & 4.97\\
& Claude-4.1-Opus  & 22.4\% & 1.361 & 4.91 & 4.98 & 4.94\\
& GPT-5            & 23.2\% & 1.330 & 4.96 & 5.00 & 4.98\\
& DeepSeek-V3      & 22.6\% & 1.413 & 4.97 & 5.00 & 4.99\\
\midrule
\multirow{4}{*}{\shortstack[l]{Prompt-based\\Compression}}
& Gemini-2.5-Pro   & 49.7\% & 0.920 & 4.44 & 4.90 & 4.67\\
& Claude-4.1-Opus  & 46.1\% & 0.852 & 4.42 & 4.93 & 4.67\\
& GPT-5            & 53.6\% & 0.937 & 4.42 & 4.87 & 4.65\\
& DeepSeek-V3      & 38.4\% & 0.960 & 4.42 & 4.87 & 4.65\\
\midrule
\multirow{4}{*}{\shortstack[l]{Best-of-$N$\\Refinement}}
& Gemini-2.5-Pro   & 61.9\% & 0.900 & 4.57 & 4.57 & 4.57\\
& Claude-4.1-Opus  & 67.6\% & 0.845 & 4.51 & 4.49 & 4.50\\
& GPT-5            & 67.2\% & 0.888 & 4.71 & 4.80 & 4.75\\
& DeepSeek-V3      & 63.6\% & 0.852 & 4.56 & 4.58 & 4.57\\
\midrule
\multirow{2}{*}{Ours}
& \textsc{Homura}\textsubscript{Rubric} & 86.3\% & 0.809 & 4.91 & 4.83 & 4.87\\
& \textsc{Homura}\textsubscript{Reason} & 89.6\% & 0.846 & 4.85 & 4.93 & 4.89\\
\bottomrule
\end{tabular}
\vspace{1mm}

\raggedright
\footnotesize
Note: \textbf{IB}: In-Bounds Rate (\%). $\rho$: syllable ratio. \textbf{Acc.}: Accuracy. \textbf{Comp.}: Completeness.

\vspace{-3mm}
\end{table}

Table~\ref{tab:human_eval_acc_comp} shows that (i) without syllable constraints, all frontier LLMs achieve near-ceiling Accuracy/Completeness, serving as an unconstrained upper bound; (ii) enforcing syllable budgets by prompting substantially degrades human-rated quality, especially Accuracy; and (iii) Best-of-$N$ provides limited and unstable gains. In contrast, our \textsc{Homura} variants deliver the best overall constrained performance, approaching the unconstrained ceiling, with \textsc{Homura}\textsubscript{Rubric} slightly favoring Accuracy and \textsc{Homura}\textsubscript{Reason} favoring Completeness.

\section{Implementation Details}
\label{sec:appendix_hyper}

Our RM training framework is built on Megatron. We use the Qwen3-32B-Chat model as the initialization. The training is conducted with a batch size of 256, using a cosine learning rate scheduler with an initial learning rate of 5e-6. All models are trained on 64  Huawei's Ascend 910B NPUs.

Our RL training framework is based on the Verl framework. We use the Qwen3-8B-Chat model as the initialization for RL training. During training, we configure a batch size of 16 and perform 16 rollouts per prompt using the GRPO algorithm. The learning rate is initialized at 1e-8, and a cosine scheduler with warm-up is applied toward the final iteration. Sampling is conducted with a temperature of 1.0, and the maximum generation length is limited to 1,024 tokens. The KL penalty coefficient $\beta$ is set to 0, effectively removing the KL constraint relative to the reference policy. 
The PPO clipping range $\epsilon$ is fixed at 0.2. All models are trained for one epoch using 8 NVIDIA H800 80G GPUs.
Other detail training hyperparameters are listed in Table \ref{tab:hyperparams}.

Our training set consists of 70K instances drawn from the same data sources as the Sand-Glass benchmark, and is constructed following the same procedure.

\begin{table}[t]
\centering
\normalsize
\setlength{\tabcolsep}{20pt}
\caption{Key hyperparameters used in training \textsc{Homura}.}
\begin{tabular}{c c}
\toprule
\textbf{Hyperparameter} & \textbf{Value} \\
\midrule
$\alpha_1 $ & 0.4  \\
$\alpha_2 $ & 0.5  \\
$\lambda_{len} $ & 0.5 \\
$\lambda_{qual} $ & 0.5  \\
$k $ & 300  \\
$\tau_{min} $ & 0  \\
$\tau_{max} $ & 0.8  \\

\bottomrule
\end{tabular}
\label{tab:hyperparams}
\end{table}

\section{Ablation Study on Reward Design}
\label{sec:appendix_reward_ablation}

This appendix provides additional ablations on the reward design of \textsc{Homura}. 
We focus on two key components: 
(i) the dynamic syllable-ratio reward, which provides the temporal compliance signal, and 
(ii) the GenRM-based quality reward, which guides semantic preservation under compression. 
These analyses examine whether the proposed reward design improves optimization stability and whether a compression-oriented reward model provides better guidance than a generic external evaluator.

\subsection{Effectiveness of Dynamic Syllable-ratio Reward}
\label{sec:appendix_dynamic_reward}

In Section~\ref{sec:dynamic_reward}, we hypothesized that rigid syllable-ratio bounds are susceptible to discretization noise, particularly in shorter utterances. 
To validate the necessity of our dynamic relaxation mechanism, we conduct an ablation study comparing our \textit{Dynamic Syllable-ratio Reward} against a static baseline on the Zh$\to$En translation task. 
Both methods share the same target objective $\mathcal{B}_L \in [0.8, 0.9]$, but the static version enforces fixed boundaries regardless of source length.

\begin{figure}[h]
    \centering
    \includegraphics[width=\linewidth]{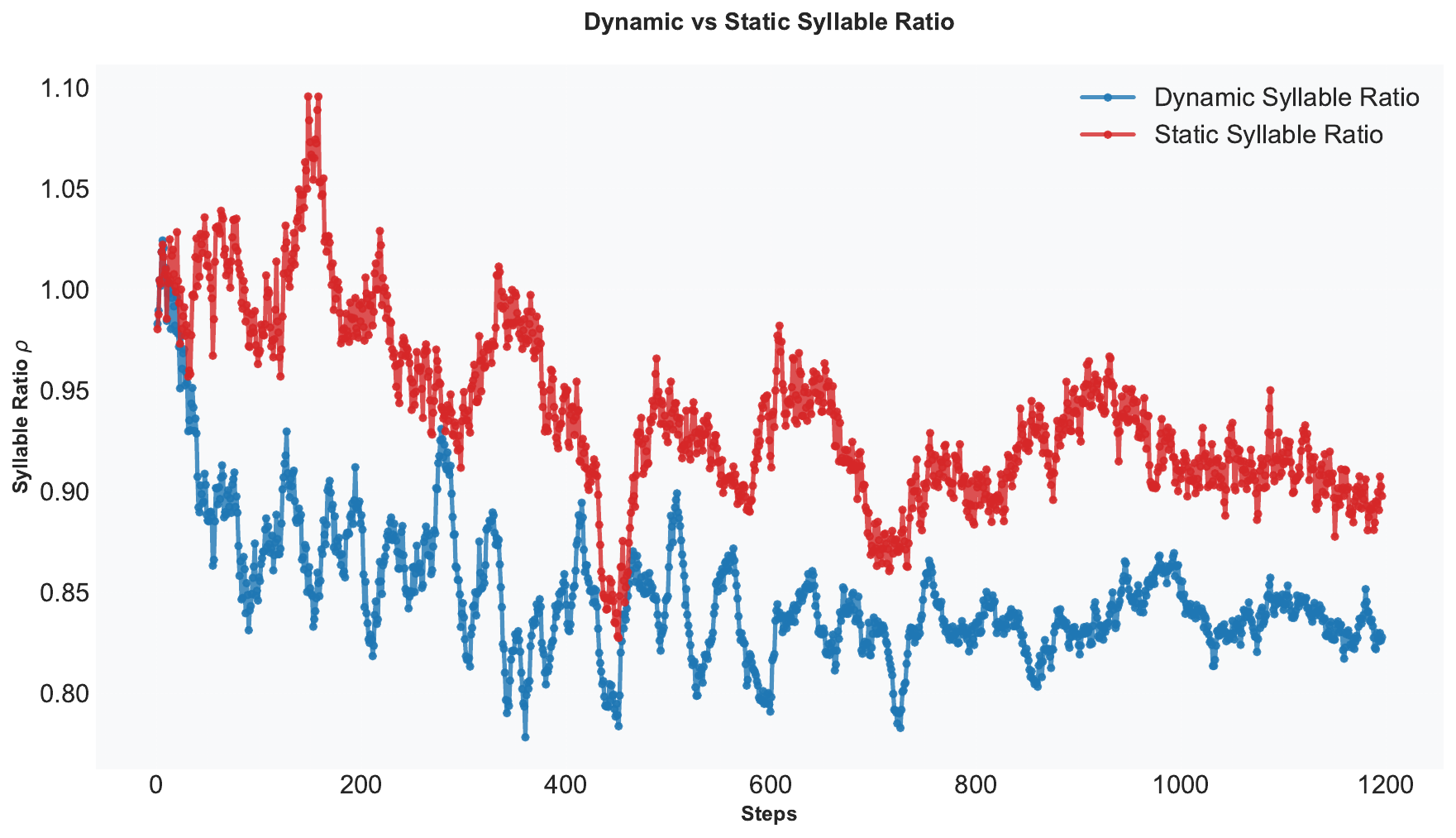}
    \caption{Training dynamics of syllable ratios during policy optimization on Zh$\to$En. 
    The blue curve represents our dynamic bound relaxation, while the red curve represents static fixed bounds. 
    Values indicate the mean syllable ratio $\rho$ across training steps.}
    \label{fig:reward_dynamics}
\end{figure}

As shown in Figure~\ref{fig:reward_dynamics}, the dynamic reward demonstrates significantly smoother optimization and faster convergence into the target interval $[0.8, 0.9]$. 
In contrast, the static reward exhibits violent oscillations, especially in early training phases. 
This instability confirms that hard rewards based on syllable counts are overly sensitive to small variations in short segments: a single-syllable difference in a short utterance can cause a disproportionate jump in the ratio, producing inconsistent learning signals for the RL agent.

By adaptively relaxing the lower bound for shorter sequences via $\gamma(x)$, our method provides a more continuous and achievable reward landscape. 
This prevents the model from being penalized for unavoidable linguistic granularity and leads to more efficient learning of the desired compression behavior. 
These findings justify the use of length-aware dynamic bounds in \textsc{Homura} to ensure stable policy refinement across diverse utterance lengths.

\subsection{Ablation of GenRM Design}
\label{sec:appendix_genrm_ablation}

We further study the design of the quality reward within the GenRM framework. 
In particular, we compare our self-trained GenRM with an external LLM-as-RM baseline. 
In this ablation, the external reward model fully replaces GenRM, while all other components of the compression-oriented reinforcement learning pipeline are kept identical.

\paragraph{External LLM-as-RM Setup.}
For the external reward model baseline, we adopt an LLM-as-RM setup using \texttt{DeepSeek-V3}. 
Given a source sentence and its translation, the model is prompted to perform a quality assessment. 
Specifically, it evaluates whether the translation 
(i) preserves the core semantic content of the source sentence under the target syllable constraint, and 
(ii) remains fluent and grammatical. 
The model outputs a binary acceptability judgment, which is directly used as the reward signal during policy optimization. 
The prompt for the external LLM-as-RM baseline is shown below.

\begin{tcolorbox}[
  enhanced,
  breakable,
  colback=white,
  colframe=black,
  boxrule=0.6pt,
  left=6pt,right=6pt,top=6pt,bottom=6pt,
  width=\columnwidth
]
\textbf{Prompt for External LLM-as-RM}

\vspace{0.5em}

You are a translation quality reward model.

Given contextual information, a source sentence, and its translation, you will perform step-by-step reasoning to assess translation quality. Your evaluation process must strictly follow the sequence below.

\textbf{Step 1: Back-Translation} \\
Translate the given translation back into the source language.

\textbf{Step 2: Semantic Consistency Assessment} \\
Compare the semantic consistency between the original source text and the back-translated text.
\begin{itemize}
  \item If the meanings are consistent or highly consistent, proceed to Step~3.
  \item If they are inconsistent, skip Step~3 and assign a final score of~0.
\end{itemize}

\textbf{Step 3: Translation Quality Assessment} \\
Evaluate whether the translation is of high quality based on factors such as fluency, grammatical correctness, and cultural appropriateness.

\vspace{0.5em}

Your final output must include a clear reasoning process and a reward score:
\begin{itemize}
  \item If semantic consistency is satisfied and the translation quality is high,
        set \texttt{score = 1}.
  \item Otherwise, set \texttt{score = 0}.
\end{itemize}

\vspace{0.5em}

\textbf{Output Format} \\
The output must strictly follow the format below. Do not add or omit any fields:
\begin{quote}\ttfamily
\noindent
\{\\
"reasoning": "<reasoning process>", \\
"score": 0 or 1 \\
\}
\end{quote}

\textbf{Input Format} \\
Context: \texttt{\{context\}} \\
Source text: \texttt{\{current\_text\}} \\
Translation: \texttt{\{translated\_text\}}
\end{tcolorbox}

\paragraph{Results and Analysis.}
Table~\ref{tab:genrm_ablation_study} compares compression translation models optimized with different reward model designs. 
The model trained with the self-trained GenRM consistently outperforms its counterpart using an external LLM-as-RM baseline, demonstrating that a reward model trained specifically for compression-oriented translation provides more effective optimization guidance than a generic prompt-based evaluator.

\begin{table}[ht]
    \centering
    \caption{Effect of GenRM choice on compressed translation under \textsc{Homura}\textsubscript{Reason}. 
    Results are averaged across all language pairs.}
    \label{tab:genrm_ablation_study}
    \small
    \setlength{\tabcolsep}{6pt}
    \renewcommand{\arraystretch}{1.1}
    \begin{tabular}{lccc}
        \toprule
        \textbf{Model Variant} & \textbf{CometKiwi} & \textbf{BT-CERR} & \textbf{BLEU-$\rho$} \\
        \midrule
        \textsc{Homura}\textsubscript{Reason} & 0.636 & 0.881 & 0.335 \\
        \midrule
        \textsc{Homura}\textsubscript{Reason} w/ External RM & 0.628 & 0.879 & 0.330 \\
        \bottomrule
    \end{tabular}
\end{table}

The self-trained GenRM achieves better CometKiwi, BT-CERR, and BLEU-$\rho$ scores than the external LLM-as-RM variant. 
This suggests that task-specific reward modeling is beneficial for compression translation, where the evaluator must judge not only general translation quality but also whether the model preserves essential meaning under strict syllable budgets. 
Moreover, GenRM is only used during training and does not introduce additional inference-time cost, preserving the single-pass decoding advantage of \textsc{Homura}.

\section{Scaling to Larger Model.}
\label{sec:appendix_scale}

\begin{table}[h]
\centering
\caption{Scaling Analysis of \textsc{Homura} (\text{Zh}$\rightarrow$\text{En}). Comparison between 8B and 32B backbones at $\mathcal{B}_L \in [0.8, 0.9]$.}
\label{tab:scaling_Homura}
\small
\setlength{\tabcolsep}{6pt}
\renewcommand{\arraystretch}{1.15}
\begin{tabular}{@{}l l c c c c c@{}}
\toprule
\textbf{Scale} & \textbf{Model} & \textbf{IB} & $\rho$ & \textbf{CK} & \textbf{BC} & \textbf{B-$\rho$} \\
\midrule
8B
 & \textsc{Homura}\textsubscript{Rubric} & 86.3\% & 0.809 & 0.700 & 0.914 & 0.378 \\
 & \textsc{Homura}\textsubscript{Reason} & 89.6\% & 0.846 & 0.701 & 0.925 & 0.376 \\
\midrule
32B
 & \textsc{Homura}\textsubscript{Rubric} & 90.2\% & 0.803 & 0.740 & 0.917 & 0.388 \\
 & \textsc{Homura}\textsubscript{Reason} & 89.9\% & 0.801 & 0.740 & 0.933 & 0.385 \\
\bottomrule
\end{tabular}

\vspace{1mm}
\raggedright
\footnotesize
Note: \textbf{IB}: In-Bounds Rate (\%). $\rho$: syllable ratio. \textbf{CK}: CometKiwi. \textbf{BC}: BT-CERR. \textbf{B-$\rho$}: BLEU-$\rho$.

\end{table}

We analyze the effect of model scaling by comparing 8B and 32B variants of \textsc{Homura} under identical training settings, with results reported in Table \ref{tab:scaling_Homura}. Scaling consistently improves translation quality and semantic preservation, as shown by higher BLEU-$\rho$, Cometkiwi and BT-CERR scores. Moreover, the relative advantage of \textsc{Homura}\textsubscript{Reason} over \textsc{Homura}\textsubscript{Rubric} is preserved at larger scale, suggesting that the proposed reward design scales robustly with model capacity.

\subsection{Ablation Study on Reward Design}
\label{sec:reward-ablation}

We further analyze the two key components of \textsc{Homura}'s reward design: 
the dynamic syllable-ratio reward for temporal compliance and the GenRM-based quality reward for semantic preservation. 
These ablations aim to answer two questions: 
(i) whether dynamic length constraints are more effective than static bounds under syllable-level granularity, and 
(ii) whether the design of the quality reward affects the compression--fidelity trade-off.

\paragraph{Effectiveness of Dynamic Syllable-ratio Reward.}
We first examine the importance of using dynamic syllable-ratio bounds. 
A static interval applies the same lower and upper thresholds to all utterances, regardless of source length. 
However, syllable counts are discrete, and short utterances are particularly sensitive to small counting errors: 
adding or removing only one syllable can cause a large change in the ratio $\rho(x,y)$. 
As a result, static bounds may over-penalize otherwise acceptable translations for short inputs and provide unstable learning signals.

To address this issue, \textsc{Homura} relaxes the lower bound for shorter utterances while keeping the upper bound fixed, producing an adaptive interval $[L(x), R(x)]$. 
Compared with the static variant, the dynamic reward yields more stable length control and better preserves translation quality. 
This suggests that the gain of \textsc{Homura} does not come from merely imposing a rigid length penalty, but from providing a smoother and length-aware optimization signal that accounts for syllable-level discreteness.

\paragraph{Ablation of GenRM Design.}
We then study the design of the quality reward. 
In addition to the rubric-based reward, \textsc{Homura} explores a reason-based generative reward model, GenRM, which jointly assesses whether the compressed translation preserves the source meaning and remains linguistically well-formed. 
To isolate the contribution of GenRM, we compare it with alternative quality reward designs, including the rubric-based reward and an LLM-as-RM variant.

The results show that GenRM improves semantic preservation under strict length constraints, especially when compression requires selecting which information should be retained. 
Unlike simple scalar rubric scores, GenRM explicitly reasons about back-translation consistency and translation quality before producing the final reward decision. 
This makes the reward less likely to favor overly short but semantically incomplete outputs. 
Importantly, GenRM is only used during training and does not introduce additional inference-time cost. 
Therefore, \textsc{Homura} retains the single-pass decoding advantage over Best-of-$N$ and translate-then-rewrite pipelines while benefiting from a stronger quality signal during RL training.

\paragraph{Discussion.}
Taken together, these ablations confirm that both sides of the reward are necessary. 
The dynamic syllable-ratio reward improves temporal compliance by stabilizing the length-control signal, while GenRM improves semantic robustness by discouraging meaning drift under compression. 
Their combination enables \textsc{Homura} to operate at a better compression--fidelity trade-off: translations remain within the target syllable budget while preserving the core predicate--argument information of the source.

\section{Examples of Structured CoT Reasoning for GenRM}
\label{sec_genrm_example_appendix}
This appendix provides illustrative examples of the structured CoT reasoning produced by GenRM.
\vspace{4mm}
\begin{tcolorbox}[
  enhanced,
  breakable,
  colback=white,
  colframe=black,
  boxrule=0.6pt,
  left=6pt,right=6pt,top=6pt,bottom=6pt,
  width=\columnwidth
]
\textbf{Structured CoT Example \#1}

\par\smallskip
\textbf{Source sentence (x):}\\
{\small 我曾经在这看到过几艘陌生的船。}

\par\smallskip
\textbf{Translation hypothesis (y):}\\
I have seen a few strange ships here before.

\par\smallskip
\textbf{COT process:}
\begin{enumerate}
  \item \textbf{Back-translation:}
  The translation ``I have seen a few strange ships here before.'' is back-translated into Chinese.
  The back-translation result is {\small ``我以前曾在这里见过几艘奇怪的船。''}
  (I have previously seen several strange ships here).
  Alternatively, it can also be {\small ``我曾经在这里看到过几艘陌生的船。''}
  (I once saw several unfamiliar ships here).

  \item \textbf{Semantic consistency judgment:}
  Compare the source sentence {\small ``我曾经在这看到过几艘陌生的船。''}
  (I once saw several unfamiliar ships here) with the back-translated content
  {\small ``我以前曾在这里见过几艘奇怪的船。''}
  (I have previously seen several strange ships here).
  The two are completely consistent in core semantics, both expressing the fact of
  {\small ``过去在某地看到过不熟悉的船只''} (having seen unfamiliar ships at a certain place in the past).
  {\small ``曾经''} (once / before) corresponds to ``before,''
  {\small ``在这''} (here) corresponds to ``here,''
  {\small ``看到过''} (have seen) corresponds to ``have seen,''
  {\small ``几艘''} (a few) corresponds to ``a few,''
  {\small ``陌生的''} (unfamiliar / strange) corresponds to ``strange,''
  and {\small ``船''} (ships) corresponds to ``ships.''
  The semantic consistency is very high.

  \item \textbf{Translation quality judgment:}
  The translation ``I have seen a few strange ships here before.'' is grammatically correct,
  using the present perfect tense to express a past experience, which conforms to English usage.
  The expression is fluent and natural, with accurate word choice. Therefore, the translation quality is high.
\end{enumerate}

Based on the above, the semantics are consistent and the translation quality is high, resulting in a final score of 1.

\par\smallskip
\textbf{Score:}\\
1
\end{tcolorbox}

\begin{tcolorbox}[
  enhanced,
  breakable,
  colback=white,
  colframe=black,
  boxrule=0.6pt,
  left=6pt,right=6pt,top=6pt,bottom=6pt,
  width=\columnwidth
]
\textbf{Structured CoT Example \#2}

\par\smallskip
\textbf{Source sentence (x):}\\
{\small 所以天命人只有超度了金池长老的冤魂}

\par\smallskip
\textbf{Translation hypothesis (y):}\\
So the Mandate Person can only rescue the lingering soul of Jinchi Elder

\par\smallskip
\textbf{COT process:}
\begin{enumerate}
  \item \textbf{Back-translation:}
  The translation ``So the Mandate Person can only rescue the lingering soul of Jinchi Elder''
  is back-translated into Chinese as
  {\small ``所以天命人只能解救金池长老逗留的灵魂。''}
  (So the Mandate Person can only rescue the lingering soul of Elder Jinchi).

  \item \textbf{Semantic consistency judgment:}
  The key terms in the source text are {\small ``超度''} (perform salvation rites / help a soul attain deliverance)
  and {\small ``冤魂''} (wronged or vengeful spirit).
  {\small ``超度''} is a term with strong religious and cultural connotations, referring to helping the deceased soul
  attain liberation and enter reincarnation or the afterlife through rituals.
  However, the word ``rescue'' in the translation is overly general and fails to convey the deeper religious meaning
  of {\small ``超度''}.
  {\small ``冤魂''} refers to a ghost that died unjustly and carries strong resentment, whereas ``lingering soul''
  {\small (逗留的灵魂)} only expresses the idea of lingering or wandering, losing the core semantic component of
  {\small ``冤''} (wronged / resentful).
  Therefore, the back-translated content is semantically inconsistent with the original text.

  \item \textbf{Translation quality judgment:}
  Due to the semantic inconsistency, this step is skipped.
\end{enumerate}

Based on the above analysis, the semantics are inconsistent, and the final score is 0.

\par\smallskip
\textbf{Score:}\\
0
\end{tcolorbox}

\section{Prompt for Linguistic well-formedness Reward.}
\label{sec:prompt_fluency}

This appendix provides the training prompt used for the linguistic well-formedness reward.

\begin{tcolorbox}[
  enhanced,
  colback=white,
  colframe=black,
  boxrule=0.6pt,
  left=6pt,right=6pt,top=6pt,bottom=6pt,
  width=\columnwidth,
  breakable
]
\textbf{Training Prompt}

You are an expert in \{target\_lang\} translation.
Given context \texttt{<context>} for background reference.
Here is a text \texttt{<text>} from the context and its corresponding
\{target\_lang\} translation \texttt{<translation>}.
You need to determine whether \texttt{<translation>} is a qualified
\{target\_lang\} translation.

\medskip
\textbf{Requirements:} Consider only the following criteria; if any are
not met, the score is 0:
\begin{enumerate}
  \item \textbf{Grammaticality:}
  No obvious grammatical or syntactic errors, consistency errors
  (e.g., subject--verb agreement, tense, case, gender/number),
  collocation errors, or linguistic defects caused by punctuation or spelling.
  \item \textbf{Readability \& Coherence:}
  The sentence structure is complete, logically coherent, and naturally
  connected, making it smooth and easy to understand; there should be no
  obvious fragmentation, garbled text, repetitive stacking, or malformed
  sentences.
\end{enumerate}

\medskip
\textbf{Input:}

\texttt{<context>}: \{context\}

\texttt{<text>}: \{text\}

\texttt{<translation>}: \{translation\}

\medskip
\textbf{Output Requirements:}
\begin{itemize}
  \item If \texttt{<translation>} satisfies both criteria above, return
  \texttt{<<1>>}.
  \item Otherwise, return \texttt{<<0>>}.
  \item Please output \texttt{<<0>>} or \texttt{<<1>>} directly; do not
  output any explanation.
\end{itemize}

\textbf{Output:}
\end{tcolorbox}

\section{Case Study}
\label{sec:appendix_cases}

Table~\ref{tab:case_study} presents two representative $\text{Zh}{\rightarrow}\text{En}$ cases to qualitatively compare \textsc{Homura} against three baseline strategies built on top of a frontier LLM (Claude-4.1-Opus).
For each case, we report: (1)~\textbf{Direct}---the unconstrained translation produced by the LLM, which serves as an upper bound on quality but ignores the syllable budget; (2)~\textbf{Truncated}---the same raw translation hard-cut at the target syllable count, simulating naïve post-hoc length control; and (3)~\textbf{Best-of-$N$ Refinement} ($k{=}4$)---four translations generated at progressively increasing compression levels via prompting, from which the candidate closest to the target interval $[0.8, 0.9]$ is selected. We present all four candidates ($t_1$--$t_4$) to visualize the coverage of the ratio space. Finally, we show the output of \textsc{Homura} under the same target ratio.

\paragraph{Observation 1: The Discrete Coverage Gap of Best-of-$N$ Refinement.}
Across both cases, the four Best-of-$N$ candidates exhibit a characteristic \emph{cliff pattern}: adjacent candidates jump across the target interval $[0.8, 0.9]$ without landing inside it.
In Case~1, the ratio drops from $t_2{=}1.45$ to $t_3{=}0.93$, and further to $t_4{=}0.55$;
in Case~2, it falls from $t_2{=}1.16$ directly to $t_3{=}0.78$.
This confirms that prompting a frontier LLM for a fixed number of compression variants cannot reliably cover the continuous ratio space---the model's coarse-grained ``short vs.\ long'' heuristic produces large gaps precisely in the moderate-compression regime where dubbing applications operate.
\textsc{Homura}, by contrast, achieves fine-grained syllable awareness during decoding and hits $\rho{=}0.83$ and $0.84$ in a single pass.

\paragraph{Observation 2: Truncation Destroys Sentence Integrity.}
When the Direct baseline over-translates (Case~1: $\rho{=}1.24$; Case~2: $\rho{=}1.22$), hard truncation to the syllable budget produces syntactically broken output---``\textit{boldly experience this\ldots}'' loses its object, and ``\textit{continued the\ldots}'' leaves a dangling determiner.
These fragments may achieve the target ratio numerically but are unusable for dubbing, where every utterance must be a self-contained, prosodically complete phrase.
\textsc{Homura} avoids this entirely: its outputs are natively generated to the target length, preserving both grammatical completeness and pragmatic tone (e.g., the encouraging ``\textit{No worries! Just dive right in}'' in Case~1 and the poetic ``\textit{the ice tale Chinese write, for a thousand years}'' in Case~2).

\paragraph{Observation 3: Register-Aware Compression Strategies.}
Beyond ratio accuracy, \textsc{Homura} adapts its compression strategy to the register of the source.
For the colloquial, exhortative Case~1, it employs informal contractions and exclamatives (``\textit{Haven't read\ldots? No worries!}'') that faithfully reproduce the casual, inviting tone of the Chinese source.
For the literary, parallel-structured Case~2, it opts for noun-phrase compaction (``\textit{hunting on wood across snow peaks}'') and appositive restructuring (``\textit{the ice tale Chinese write}''), achieving compression through poetic density rather than information loss.
This register sensitivity suggests that the phoneme-indexed reward signal trains the model to internalize not just \emph{how many} syllables to produce, but \emph{which linguistic register} best accommodates the budget.

\begin{table*}[!ht]
\centering
\caption{Qualitative case studies on $\text{Zh}\rightarrow\text{En}$ translation with syllable-ratio control (target $\rho \in [0.8, 0.9]$).
\textit{Direct} translates without length awareness; \textit{Truncated} hard-cuts the raw output to the syllable budget;
\textit{Multi} generates four candidates at varying compression levels and selects the closest to the target interval.}
\label{tab:case_study}
\footnotesize
\setlength{\tabcolsep}{4pt}
\renewcommand{\arraystretch}{1.25}
\begin{tabularx}{\linewidth}{@{} l l X c c p{0.21\linewidth} @{}}
\toprule
\textbf{Category} & \textbf{Method} & \textbf{Output} & \textbf{Syl.} & $\boldsymbol{\rho}$ & \textbf{Observation} \\
\midrule
\multicolumn{6}{@{}l}{\textit{Case~1.}\enspace 没看过原著的朋友更不用担心，大胆去体验这个瑰奇的打工人故事吧\enspace\textit{(29\,syl.)}} \\[3pt]
\multirow{2}{*}{Direct}
  & Raw    & Friends who haven't read the original work, don't worry even more. Go ahead and boldly experience this wonderful story of a working man. & 36 & 1.24 & Fluent; ignores syllable budget. \\
  & Trunc. & Friends who haven't read the original work, don't worry even more. Go ahead and boldly experience this{\,\ldots} & 26 & 0.90 & In range but object missing; broken. \\
\cmidrule(lr){1-6}
\multirow{4}{*}{\shortstack[l]{Multi\\($k{=}4$)}}
  & $t_1$ & For those of you who haven't read the original novel, there's even less to worry about; feel free to boldly experience this wonderfully peculiar story of a working person. & 47 & 1.62 & Heavily over-expanded. \\
  & $t_2$ & Friends who haven't read the original novel have even less to worry about; boldly go and experience this fantastic story of a working-class individual. & 42 & 1.45 & Still far above target. \\
  & $t_3$ (sel.) & If you haven't read the book, don't worry; go experience this amazing story of a working person. & 27 & 0.93 & Closest candidate; overshoots. \\
  & $t_4$ & Unread the book? No worries, experience this worker's tale. & 16 & 0.55 & Over-compressed; target skipped. \\
\cmidrule(lr){1-6}
\textsc{Homura} & $\rho{\approx}0.83$ & Haven't read the book? No worries! Just dive right in and enjoy this fantastic working-man's tale! & 24 & \textbf{0.83} & \textbf{On target}; encouraging tone preserved. \\
\midrule
\multicolumn{6}{@{}l}{\textit{Case~2.}\enspace 从雪山踏木狩猎而行，再到世界舞台上的比拼，冰雪的故事，中国人续写了千年\enspace\textit{(32\,syl.)}} \\[3pt]
\multirow{2}{*}{Direct}
  & Raw    & From hunting in the snowy mountains with wooden skis to competing on the world stage, the Chinese have continued the story of ice and snow for thousands of years. & 39 & 1.22 & Over-translation; verbose. \\
  & Trunc. & From hunting in the snowy mountains with wooden skis to competing on the world stage, the Chinese have continued the{\,\ldots} & 28 & 0.88 & Broken at ``continued the''. \\
\cmidrule(lr){1-6}
\multirow{4}{*}{\shortstack[l]{Multi\\($k{=}4$)}}
  & $t_1$ & From hunting on snow-covered mountains with wooden skis to competing on the global stage, the story of ice and snow has been continued by Chinese people for thousands of years. & 44 & 1.38 & Heavily over-expanded. \\
  & $t_2$ & From mountain hunting with wooden skis to global competitions, Chinese people have continued the story of ice and snow for a millennium. & 37 & 1.16 & Still above target. \\
  & $t_3$ (sel.) & From hunting on snowy mountains to world competitions, China's ice and snow story spans millennia. & 25 & 0.78 & Below target; interval skipped. \\
  & $t_4$ & China's ice and snow story, from ancient hunts to modern competition, spans millennia. & 23 & 0.72 & Further compressed; target missed. \\
\cmidrule(lr){1-6}
\textsc{Homura} & $\rho{\approx}0.84$ & From hunting on wood across snow peaks, to competing on the world stage---the ice tale Chinese write, for a thousand years. & 27 & \textbf{0.84} & \textbf{On target}; poetic compression. \\
\bottomrule
\end{tabularx}
\end{table*}

\section{Detailed Breakdown of Expansion Metrics}
\label{sec:appendix_noraml_roundtrip_metrics}

\begin{figure*}[t]
\centering
\includegraphics[width=0.95\textwidth]{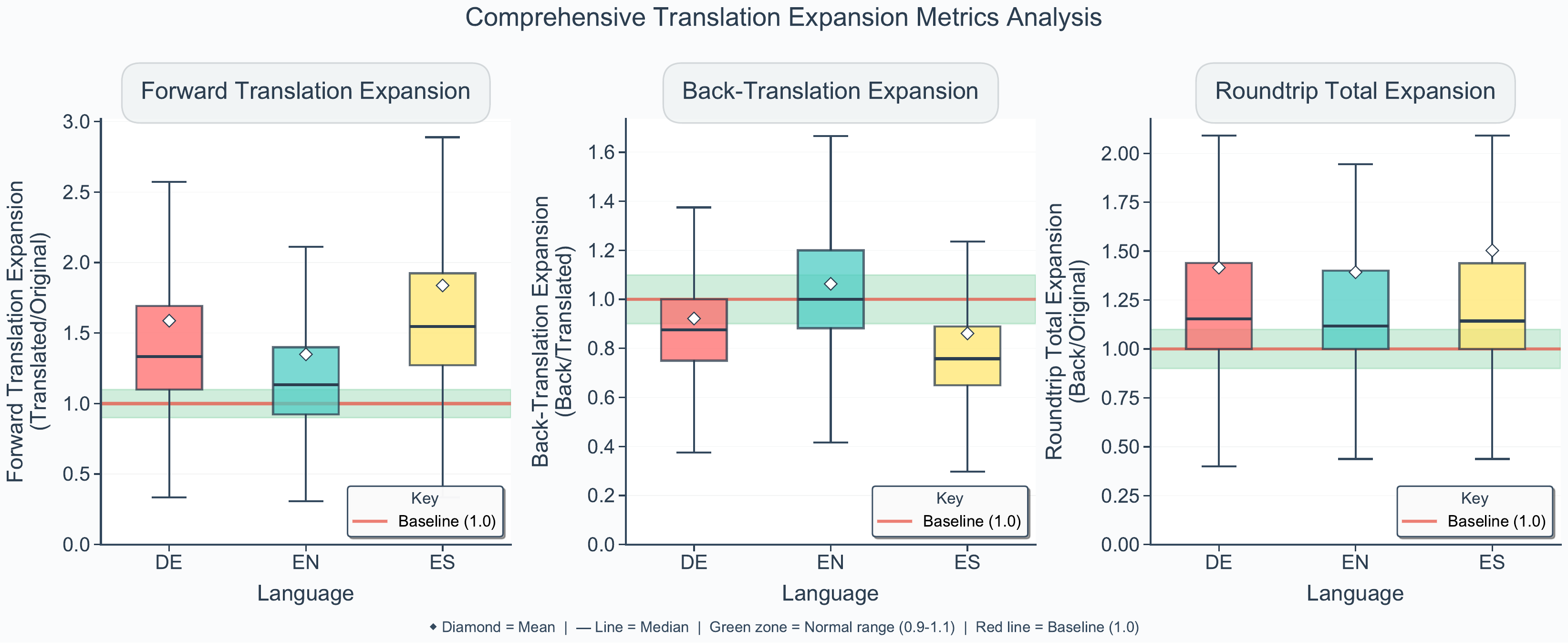}
\vspace{-2mm}
\caption{Statistical distribution of translation expansion metrics across DE, EN, and ES. The boxplots illustrate the variations in Forward, Back, and Roundtrip expansion ratios, with diamonds indicating means.}
\label{fig:boxplot_three_metrics}
\vspace{-3mm}
\end{figure*}

To provide a comprehensive view of the verbosity bias identified in Section~\ref{sec:pilot}, we present a fine-grained statistical breakdown of expansion metrics across various frontier LLMs and language pairs. As summarized in Table~\ref{tab:detailed_metrics}, the \textit{Roundtrip Expansion Ratio} ($R_{rtp}$) consistently exceeds the unity baseline across all tested models and languages, with a high percentage of segments ($>60\%$) exhibiting systemic inflation. 

The distribution shifts visualized in Figure~\ref{fig:boxplot_three_metrics} further confirm that this bias is not merely a result of specific difficult cases but represents a structural characteristic of current LLM-based translation. 

\begin{table*}[!th]
\centering
\caption{Detailed Breakdown of Expansion Metrics across Models and Language Pairs. $R_{fwd}$, $R_{bwd}$, and $R_{rtp}$ denote Forward, Backward (Back-translation), and Roundtrip Expansion Ratios, respectively. Values are presented as Mean and Standard Deviation.}
\label{tab:detailed_metrics}
\small
\begin{tabular*}{\textwidth}{@{\extracolsep{\fill}}llcccc@{}}
\toprule
\textbf{Language} & \textbf{Model} & \textbf{$R_{fwd}$ (Mean $\pm$ Std)} & \textbf{$R_{bwd}$ (Mean $\pm$ Std)} & \textbf{$R_{rtp}$ (Mean $\pm$ Std)} & \textbf{$R_{rtp} > 1$ (\%)} \\ \midrule

\multirow{5}{*}{\textbf{De}} 
 & \texttt{claude-4.1-opus} & $1.58 \pm 0.96$ & $0.95 \pm 1.54$ & $1.39 \pm 0.99$ & 66.2 \\
 & \texttt{deepseek-v3} & $1.67 \pm 1.37$ & $0.91 \pm 0.29$ & $1.54 \pm 1.89$ & 63.8 \\
 & \texttt{gemini-2.5-pro} & $1.51 \pm 0.88$ & $0.96 \pm 0.44$ & $1.35 \pm 0.84$ & 59.0 \\
 & \texttt{gpt-5} & $1.63 \pm 0.96$ & $0.90 \pm 0.19$ & $1.43 \pm 0.87$ & 71.8 \\ \midrule

\multirow{5}{*}{\textbf{En}} 
 & \texttt{claude-4.1-opus} & $1.36 \pm 1.13$ & $1.07 \pm 0.55$ & $1.41 \pm 1.32$ & 64.8 \\
 & \texttt{deepseek-v3} & $1.41 \pm 1.15$ & $1.05 \pm 0.48$ & $1.47 \pm 1.71$ & 59.1 \\
 & \texttt{gemini-2.5-pro} & $1.29 \pm 0.81$ & $1.07 \pm 0.25$ & $1.33 \pm 0.80$ & 61.0 \\
 & \texttt{gpt-5} & $1.33 \pm 0.81$ & $1.07 \pm 0.23$ & $1.36 \pm 0.80$ & 65.9 \\ \midrule

\multirow{5}{*}{\textbf{Es}} 
 & \texttt{claude-4.1-opus} & $1.81 \pm 1.05$ & $0.90 \pm 2.48$ & $1.41 \pm 1.19$ & 66.9 \\
 & \texttt{deepseek-v3} & $1.99 \pm 1.83$ & $1.01 \pm 1.12$ & $1.99 \pm 2.90$ & 64.0 \\
 & \texttt{gemini-2.5-pro} & $1.72 \pm 0.99$ & $0.83 \pm 0.36$ & $1.35 \pm 0.87$ & 60.4 \\
 & \texttt{gpt-5} & $1.86 \pm 1.09$ & $0.78 \pm 0.18$ & $1.40 \pm 0.83$ & 68.5 \\ \bottomrule
\end{tabular*}
\end{table*}

\section{Declaration of AI Assistance}
To enhance development efficiency, GitHub Copilot was utilized as a coding assistant in this study. Specifically, it facilitated the implementation of data preprocessing modules, the execution of statistical analyses, and the rendering of complex experimental figures. The core algorithmic design, reinforcement learning frameworks, and the interpretation of results remain the original work of the authors, who maintain full responsibility for the code's accuracy.

\section{Generalization to General Translation}
\label{sec:generalization_wmt}

\paragraph{Generalization to Out-of-Distribution Benchmarks.}
To examine whether our method generalizes beyond the \textsc{Sand-Glass} distribution, we conduct zero-shot evaluations on two external benchmarks. First, we evaluate on WMT24pp using models trained directly on the \textsc{Sand-Glass} training set. We compare \textsc{Homura} with strong baseline models under four settings: unconstrained generation, prompting-based constrained generation, Best-of-$N$ selection, and a translate-then-rewrite pipeline.

Second, we evaluate on \textsc{MuST-Cinema}, a subtitle-oriented speech translation benchmark. We randomly sample 1,000 examples from the English-to-German subset. To match our high-density-to-low-density setting, we first translate the English source texts into Chinese using DeepSeek-V3.2, and then evaluate models on the resulting Zh$\rightarrow$De task. We compare \textsc{Homura} with \texttt{claude-4.1-opus} prompted with explicit syllable-ratio constraints. Importantly, \textsc{Homura} is evaluated zero-shot on this dataset, without any additional training or fine-tuning.

\begin{table}[t]
\centering
\small
\setlength{\tabcolsep}{6pt}
\caption{
Zero-shot generalization results on WMT24pp. All systems are evaluated without additional adaptation.
}
\begin{tabular}{llccccc}
\toprule
\textbf{Category} & \textbf{Model} & \textbf{IB} & $\rho$ & \textbf{CK} & \textbf{BC} & \textbf{B-$\rho$} \\
\midrule
\multirow{4}{*}{\shortstack[l]{Unconstrained\\LLMs}}
& Claude-4.1-Opus  & 33.3\% & 1.021 & 0.842 & 1.000 & 0.586 \\
& DeepSeek-V3       & 30.5\% & 1.073 & 0.833 & 0.969 & 0.548 \\
& Gemini-2.5-Pro    & 34.3\% & 0.998 & 0.837 & 0.989 & 0.599 \\
& GPT-5             & 30.9\% & 1.034 & 0.843 & 1.000 & 0.600 \\
\midrule
\multirow{4}{*}{\shortstack[l]{Prompt-based\\Compression}}
& Claude-4.1-Opus  & 62.6\% & 0.867 & 0.734 & 1.000 & 0.608 \\
& DeepSeek-V3       & 46.4\% & 1.094 & 0.773 & 0.979 & 0.540 \\
& Gemini-2.5-Pro    & 59.6\% & 0.803 & 0.720 & 0.989 & 0.606 \\
& GPT-5             & 84.8\% & 0.888 & 0.829 & 0.989 & 0.611 \\
\midrule
\multirow{4}{*}{\shortstack[l]{Best-of-$N$\\Refinement}}
& Claude-4.1-Opus  & 63.6\% & 0.852 & 0.720 & 1.000 & 0.619 \\
& DeepSeek-V3       & 61.2\% & 0.874 & 0.730 & 1.000 & 0.578 \\
& Gemini-2.5-Pro    & 77.7\% & 0.845 & 0.733 & 1.000 & 0.548 \\
& GPT-5             & 89.6\% & 0.847 & 0.797 & 1.000 & 0.605 \\
\midrule
\multirow{4}{*}{\shortstack[l]{Pipeline-based\\Post-editing}}
& Claude-4.1-Opus  & 31.3\% & 0.707 & 0.814 & 0.989 & 0.616 \\
& DeepSeek-V3       & 28.4\% & 0.704 & 0.804 & 0.969 & 0.609 \\
& Gemini-2.5-Pro    & 54.5\% & 0.792 & 0.824 & 0.989 & 0.622 \\
& GPT-5             & 77.3\% & 0.804 & 0.816 & 0.989 & 0.594 \\
\midrule
Ours
& \textsc{Homura}            & 86.8\% & 0.803 & 0.803 & 0.990 & 0.605 \\
\bottomrule
\end{tabular}

\vspace{1mm}
\raggedright
\footnotesize
Note: \textbf{IB}: In-Bounds Rate (\%). $\rho$: syllable ratio. \textbf{CK}: CometKiwi. \textbf{BC}: BT-CERR. \textbf{B-$\rho$}: BLEU-$\rho$.

\label{tab:wmt_generalization}
\end{table}

\begin{table}[t]
    \centering
    \caption{Zero-shot results on \textsc{MuST-Cinema} Zh$\rightarrow$De.}
    \label{tab:must_cinema}
    \small
    \setlength{\tabcolsep}{6.5pt}
    \renewcommand{\arraystretch}{1.05}

    \begin{tabular}{lccccc}
        \toprule
        \textbf{Model} & \textbf{IB} & $\rho$ & \textbf{CK} & \textbf{BC} & \textbf{B-$\rho$} \\
        \midrule
        \texttt{Claude-4.1-Opus}
        & 57.0\% & 0.999 & 0.777 & \textbf{0.976} & \textbf{0.498} \\
        \textsc{Homura}\textsubscript{Rubric}
        & 68.5\% & 0.942 & \textbf{0.781} & 0.966 & 0.475 \\
        \bottomrule
    \end{tabular}

    \vspace{1mm}
    \raggedright
    \footnotesize
    Note: \textbf{IB}: In-Bounds Rate (\%). $\rho$: syllable ratio. \textbf{CK}: CometKiwi. \textbf{BC}: BT-CERR. \textbf{B-$\rho$}: BLEU-$\rho$.
    \vspace{-3mm}
\end{table}

As shown in Table~\ref{tab:wmt_generalization}, \textsc{Homura} achieves stable zero-shot performance on WMT24pp, suggesting that the learned syllable-level control policy can transfer beyond the \textsc{Sand-Glass} distribution. Although WMT24pp differs substantially from \textsc{Sand-Glass} and contains relatively simpler translation instances, \textsc{Homura} maintains competitive translation quality while preserving effective verbosity control. In particular, \textsc{Homura} obtains a BT-CERR score of 0.990, a BLEU-$\rho$ score of 0.605, and a CometKiwi score of 0.803.

Compared with prompting-based constrained generation and Best-of-$N$ selection, \textsc{Homura} provides a more direct way to internalize syllable-ratio constraints into the model policy, avoiding repeated sampling or external selection at inference time. Pipeline-based post-editing methods can also achieve strong performance on WMT24pp, but they rely on an additional rewriting stage and may introduce extra latency and error propagation. By contrast, \textsc{Homura} performs controlled translation in a single decoding pass.

As shown in Table~\ref{tab:must_cinema}, \textsc{Homura} also transfers to \textsc{MuST-Cinema} without dataset-specific training. It achieves in-bound outputs, whereas the prompted Claude baseline exceeds the target budget. Despite using shorter outputs, \textsc{Homura} maintains competitive quality, with a slightly higher CometKiwi score and comparable BT-CERR and BLEU-$\rho$.

Overall, these results suggest that \textsc{Homura} mitigates verbosity bias without substantially compromising general translation performance. The stable transfer from \textsc{Sand-Glass} to WMT24pp and \textsc{MuST-Cinema} further indicates that \textsc{Sand-Glass} provides a challenging and practically relevant benchmark for length- and time-constrained translation.

\section{Syllables Per Second (SPS) Analysis}
\label{sec:sps}

Beyond matching the target syllable budget, a practical dubbing or subtitling system must ensure that the translated output is \emph{deliverable}---i.e., a speaker can naturally articulate it within the allotted duration. We therefore introduce \textbf{Syllables Per Second (SPS)} as a complementary metric to evaluate information density from a temporal perspective.

Leveraging the precise timestamps provided by our benchmark, we compute SPS as the ratio of the predicted syllable count to the source segment duration.
Established psycholinguistic research reports an average English syllabic rate of approximately 6.19~SPS~\citet{Pellegrino2011language}, representing the upper bound of comfortable speech delivery. Consistent with this finding, our internal human-annotated studies on live dubbing operations identify \textbf{SPS\,$<$\,6} as the critical comfort threshold for English audiences.

We evaluate all baseline categories defined before on the Zh\,$\rightarrow$\,En task and report two metrics: (1)~Average SPS across all segments, and (2)~Pass Rate, the proportion of segments satisfying SPS\,$<$\,6. Results are summarized in Table~\ref{tab:sps}.

\begin{table}[t]
\centering
\caption{SPS analysis on the Zh\,$\rightarrow$\,En task. \textbf{Avg.\ SPS} is the mean syllables per second across all segments. \textbf{Pass Rate} is the fraction of segments with SPS\,$<$\,6.}
\label{tab:sps}
\small
\setlength{\tabcolsep}{8pt}
\begin{tabular}{lcc}
\toprule
\textbf{Model} & \textbf{Avg.\ SPS} & \textbf{Pass Rate @ SPS $<$ 6} \\
\midrule
\multicolumn{3}{l}{\textit{Category 1: Unconstrained LLMs}} \\
\quad Claude-4.1-Opus   & 6.45  & 54.2\% \\
\quad DeepSeek-V3       & 6.69  & 51.7\% \\
\quad Gemini-2.5-Pro    & 6.14  & 56.7\% \\
\quad GPT-5             & 6.30  & 52.6\% \\
\midrule
\multicolumn{3}{l}{\textit{Category 2: Prompt-based Compression}} \\
\quad Claude-4.1-Opus   & 4.839 & 79.3\% \\
\quad DeepSeek-V3       & 5.404 & 72.3\% \\
\quad Gemini-2.5-Pro    & 5.247 & 71.3\% \\
\quad GPT-5             & 5.346 & 68.9\% \\
\midrule
\multicolumn{3}{l}{\textit{Category 3: Best-of-N Refinement}} \\
\quad Claude-4.1-Opus   & 4.820 & 80.6\% \\
\quad DeepSeek-V3       & 4.849 & 79.4\% \\
\quad Gemini-2.5-Pro    & 5.065 & 77.3\% \\
\quad GPT-5             & 5.042 & 76.5\% \\
\midrule
\multicolumn{3}{l}{\textit{Category 4: Pipeline Post-editing}} \\
\quad Claude-4.1-Opus   & 5.178 & 79.6\% \\
\quad DeepSeek-V3       & 5.443 & 70.5\% \\
\quad Gemini-2.5-Pro    & 4.745 & 80.3\% \\
\quad GPT-5             & 4.725 & 83.0\% \\
\midrule
\multicolumn{3}{l}{\textit{Ours}} \\
\quad \textsc{Homura} & \textbf{4.58} & \textbf{86.7\%} \\
\bottomrule
\end{tabular}
\end{table}

Several key observations emerge. First, unconstrained frontier LLMs (Category~1) consistently exceed the 6~SPS comfort threshold, with average SPS ranging from 6.14 to 6.69 and pass rates barely surpassing 50\%. This confirms a systematic \emph{verbosity bias}: when translating from a high-density language (Chinese) to a lower-density one (English), standard models produce outputs that are too long to be naturally spoken within the original duration.

Adding prompt-level syllable constraints (Category~2) substantially reduces SPS, yet pass rates remain between 69--79\%, suggesting that most models struggle to reliably internalize explicit length instructions. Best-of-N sampling (Category~3) further improves pass rates to 76--81\%, but at the cost of $N\times$ inference overhead. Pipeline post-editing (Category~4) yields competitive SPS for Gemini-2.5-Pro and GPT-5; however, the two-stage process introduces its own failure modes---Claude-4.1-Opus and DeepSeek-V3 still exhibit elevated SPS, indicating that post-hoc rewriting does not consistently respect the syllable budget.

In contrast, \textsc{Homura} achieves the lowest average SPS (4.58) and the highest pass rate (86.7\%), outperforming the best pipeline configuration by 3.7 percentage points. By learning temporal constraints as hard rewards during RL fine-tuning, \textsc{Homura} internalizes the information density trade-off directly into the generation process, eliminating the need for external post-processing or repeated sampling.

\begin{table}[htbp]
\centering
\renewcommand{\arraystretch}{1.15}
\small
\setlength{\tabcolsep}{7pt}
\begin{tabular}{lcccccc}
\toprule
\textbf{Domain} & \textbf{Samples} & \textbf{\shortstack{Avg. Chars \\ (ZH)}} & \textbf{\shortstack{Total Dur. \\ (Min)}} & \textbf{\shortstack{Avg. Dur. \\ (Sec)}} & \textbf{\shortstack{Min Dur. \\ (Sec)}} & \textbf{\shortstack{Max Dur. \\ (Sec)}} \\
\midrule
ACGN               & 200   & 12.02 & 6.37  & 1.91 & 0.56 & 4.11 \\
Film \& Television & 200   & 11.71 & 7.90  & 2.37 & 0.76 & 6.31 \\
Travel \& Tourism  & 200   & 11.52 & 9.40  & 2.82 & 0.60 & 7.52 \\
Gaming             & 200   & 12.13 & 8.93  & 2.68 & 0.92 & 6.61 \\
General Knowledge  & 200   & 12.14 & 9.73  & 2.92 & 0.64 & 9.06 \\
\midrule
\textbf{Overall}   & \textbf{1,000} & \textbf{11.90} & \textbf{42.32} & \textbf{2.54} & \textbf{0.56} & \textbf{9.06} \\
\bottomrule
\end{tabular}
\caption{Fine-grained per-domain and global descriptive statistics of the \textsc{Sand-Glass} source benchmark.}
\label{tab:domain-stats-appendix-combined}
\end{table}

\section{Domain Distribution and Fine-Grained Textual Attributes}
\label{sec:appendix_dataset_stats}
The benchmark comprises exactly 1,000 high-quality golden instances, strictly balanced with 200 instances allocated to each of the five target media domains to prevent domain-specific evaluation bias. Table~\ref{tab:domain-stats-appendix-combined} provides a fine-grained, comprehensive breakdown of the dataset's textual and temporal attributes, capturing both per-domain variations and global corpus metrics.

As indicated by the empirical properties summarized in Table~\ref{tab:domain-stats-appendix-combined}, the source segments are derived from colloquial, fast-paced digital platform content. 

At the macro level, the complete benchmark encompasses an aggregate audio footprint of \textbf{42.32 minutes} of continuous, synchronized speech. The individual segment durations span a highly dynamic and non-trivial range, from an ultra-short minimum of 0.56 seconds (frequently capturing urgent interjections or compact predicate structures in \textit{ACGN}) up to a maximum upper bound of 9.06 seconds (characterizing dense information delivery and explanatory clauses in \textit{General Knowledge}). 

The corpus exhibits a stable global mean temporal budget of 2.54 seconds per utterance tightly coupled with an average textual density of 11.90 Chinese characters. This wide yet strictly calibrated variance guarantees that constrained reinforcement learning policies optimized under the \emph{Homura} framework are systematically evaluated against diverse temporal bottlenecks, faithfully simulating the fluid pacing and volatile constraint environments of real-world media localization pipelines.

\section{Empirical Validation of Syllable as Duration Proxy}\label{app:tts_validation}

To quantitatively validate the use of syllable count as a proxy for physical speech duration under practical TTS deployment, we conduct two complementary experiments on the \textsc{Sand-Glass} benchmark.

\paragraph{Experiment 1: Correlation between Syllables and Speech Duration.}
We synthesize the English translations of a held-out subset using \textbf{9 TTS configurations} spanning diverse paradigms: cloud neural (\textsc{edge-tts} Aria, Guy, Sonia), flow-matching (\textsc{Kokoro} af\_heart, am\_michael), VITS (\textsc{Piper} lessac-medium, ryan-high), on-device transformer (\textsc{KittenTTS}), and rule-based formant (\textsc{espeak-ng}). For each synthesized segment, we measure the silence-trimmed audio duration. Table~\ref{tab:tts_corr} reports the Pearson correlation $r$ between syllable count and measured duration, as well as the coefficient of determination $R^2$ from a linear regression of duration on syllable count augmented with a punctuation-break count (periods, commas, and question marks, serving as a proxy for pause insertions).

\begin{table}[htbp]
\centering
\small
\caption{Correlation between syllable count and measured TTS duration across 9 synthesis systems.}
\begin{tabular}{lcc}
\toprule
\textbf{System} & \textbf{$r$} & \textbf{$R^2$ (+ punct.)} \\
\midrule
edge-tts / Aria (US-F) & 0.913 & 0.950 \\
edge-tts / Guy (US-M)  & 0.887 & 0.945 \\
edge-tts / Sonia (GB-F) & 0.943 & 0.948 \\
Kokoro / af\_heart (F)  & 0.975 & 0.954 \\
Kokoro / am\_michael (M) & 0.975 & 0.954 \\
Piper / lessac-medium   & 0.960 & 0.939 \\
Piper / ryan-high       & 0.963 & 0.932 \\
espeak-ng (formant)     & 0.953 & 0.944 \\
KittenTTS (15M)         & 0.973 & 0.956 \\
\bottomrule
\end{tabular}
\label{tab:tts_corr}
\end{table}

Speech duration is highly linearly correlated with syllable count in every system ($r = 0.89\text{--}0.98$). Even for markedly slower or faster voices, the relationship remains linear, reducing voice- and rate-differences to a one-scalar calibration. Cross-system residuals stem chiefly from punctuation-induced pauses, which are themselves text-predictable; adding a single punctuation-break term yields $R^2 = 0.93\text{--}0.96$ for every system.

\paragraph{Experiment 2: Real Time Budget Compliance in Spoken Speech.}
We synthesize every baseline's and \textsc{Homura}'s translations on the \textsc{Sand-Glass} (Zh$\to$En) subset with a single fixed TTS voice (\textsc{Kokoro}, the system with near-zero intercept in the above table, i.e., the most neutral duration proxy). We compare the measured, silence-trimmed spoken duration against each segment's original timestamp budget. Table~\ref{tab:tts_budget} reports the median ratio of spoken duration to the time budget, where a value of $1.0$ indicates that the speech exactly fills the allotted time slot. Each baseline row averages the four frontier systems from Table~3 within that category.

\begin{table}[htbp]
\centering
\small
\caption{Median spoken duration to time budget ratio across strategies (Zh$\to$En).}
\begin{tabular}{lc}
\toprule
\textbf{System} & \textbf{Median dur / budget} \\
\midrule
Unconstrained LLMs          & 1.32 \\
Prompt-based compression    & 1.13 \\
Best-of-$N$                 & 0.87 \\
Pipeline post-editing       & 1.10 \\
\textsc{Homura}             & \textbf{1.02} \\
\bottomrule
\end{tabular}
\label{tab:tts_budget}
\end{table}

Unconstrained models typically run $32\%$ over the time slot, while prompt-based and pipeline baselines still exceed by over $10\%$. Best-of-$N$ fits the numerical budget only by systematically undershooting ($0.87\times$), which leaves the target slot noticeably underfilled in actual speech. In contrast, \textsc{Homura}'s spoken output is essentially centered on the real time budget ($1.02\times$) in a single pass, confirming that the syllable-constrained objective effectively delivers time-constrained behavior when rendered as actual speech.

\section{Baseline Prompt Templates}\label{app:baseline_prompts}

We provide verbatim prompt templates for all baseline categories used in Table~3. Original Chinese instructions are shown with English translation in brackets.

\subsection{Category 1: Unconstrained}

\begin{verbatim}
给定上下文，请你将输入的中文音频文本内容，确保完整地翻译成{target_language}。
上文: {up_context}
下文: {down_context}
请对输入进行翻译，直接输出翻译结果。
输入: {input_text}
输出:
\end{verbatim}

\noindent (English: ``Given the context, translate the input Chinese audio text completely into \texttt{\{target\_language\}}. Preceding context: \texttt{\{up\_context\}}. Following context: \texttt{\{down\_context\}}. Translate the input and output the translation directly. Input: \texttt{\{input\_text\}}. Output:'')

\subsection{Category 2: Prompt-based Compression}

\begin{verbatim}
请你将输入的中文视频音频内容翻译成{target_language}，输入为一个json格式。

context：包含当前文案以及其上下文信息，帮助你理解，不需要翻译。
current_text：需要翻译的中文内容。
syllable：当前中文的音节数量。

翻译要求:
a. 音节个数控制限制：你需要保证翻译后的 {target_language}音节数量 / 中文音节数量 的音节比在 [{ratio_min}, {ratio_max}] 区间内，你可以通过增删翻译内容来达到这个要求。
b. 翻译完整性：请尽可能保留中文内容的完整性和语义，避免遗漏信息。

格式示例如下，你只需要对当前文案进行翻译，上下文仅为你提供参考信息:
输入: 
{
    "context": "xxxxxxxxx", 
    "current_text": "xxx", 
    "syllable": "xx"
}
输出: xxx

现在请根据上述要求完成如下视频音频内容的翻译，直接输出{target_language}翻译结果，不要进行任何解释。
输入: 
{
    "context": {context}, 
    "current_text": {current_text}, 
    "syllable": {syllable}
}
输出:
\end{verbatim}

\noindent (English: the model receives a JSON with the context, the current Chinese text, and its syllable count; it must (a) keep the target/source syllable ratio within \texttt{[\{ratio\_min\}, \{ratio\_max\}]}, adjusting content as needed, and (b) preserve the completeness and semantics of the source as much as possible; it outputs the translation directly without explanation.)

\subsection{Category 3: Best-of-N}

\begin{verbatim}
## 请你将输入的视频音频内容翻译成{target_language}配音脚本，符合native speaker的说话表达习惯，输入为一个json格式，key为序号，value为内容，一共有{x}条文本。
请你把每条文本翻译成{target_language}后输出，要求输出的每句都和输入按顺序匹配。
对于每条文本，你需要先翻译一个base版本的文本，也就是是最符合要求的版本，然后在这个版本基础上进行扩展或者缩减，以适应不同的配音时间要求。

## 格式示例如下:
请注意base_translated_text指的是你输出的base翻译结果，之后以base为基准，translated_text输出你的改写结果，这是你为了适应不同的配音时间要求而进行删减或者扩展的版本，需保证translated_text输出的翻译结果长短有显著区别。
具体来说，请确保translated_text的字数长度差异明显，并且从长至短的输出各结果：
第一个：长度最长，在base版本的基础上进行适当额外补充，比如增加修饰词和副词，不改变语义，仅增加情感表达；
第二个：base版本；
第三个：在base版本的基础上删减不必要的修饰副词，不改变语义；
第四个：仅保留base的主谓宾定语和其他关键部分，在不改变语义的基础上比第三句简短一些。
请注意，在第三句和第四句的缩减过程中，你需要确保以下三个要求：
a. 完整传递源语70%的信息，省略得当（修辞手法、四字短语、语气词、同义词及范畴词），无关键信息遗漏。
b. 内容完整性：确保句子的主要信息和逻辑关系100%完整。主从句完整性：主句和从句（非状语、非补语）必须保留主谓语。仅允许进行浓缩或代词替换。其他成分处理：宾语、补语、插入语和修饰语可以进行压缩性意译或删除。关键成分保留：数词、动词及其受者不可省略，必须完整保留。

输入:
上文: {up_context}
下文: {down_context}
输入: {input_text}

你只需要对输入进行翻译即可，不需要考虑上下文。
输出: {
"origin_text": "xxx", 
"base_translated_text": ["xxx"], 
"translated_text": 
["aaaaa", "bbbb", "ccc", "dd"]
}

现在请根据上述要求完成如下视频音频内容的翻译，输出各翻译结果，直接返回{target_language}，不要进行任何解释。
输入:
\end{verbatim}

\noindent (the model first produces a base translation, then rewrites it into four length-graded variants — expanded, base, trimmed, and maximally condensed — under explicit content-preservation rules. Among the candidates, the one whose syllable count best fits the duration constraint is selected as the final output.)

\subsection{Category 4: Pipeline Post-editing}

\paragraph{Stage 1 (Translation):}
Same as Category 1 (Unconstrained) template above.

\paragraph{Stage 2 (Length Adjustment):}

\begin{verbatim}
给定中文及其{target_language}翻译，请将翻译进行音节维度的长度调整。

text：原文
translation：翻译

要求音节个数控制限制：你需要保证翻译后的 {target_lang_name}音节数量 / 中文音节数量 在 {range} 区间内，你可以通过增删翻译内容来达到这个要求。

请根据上述要求进行翻译的长度调整，直接输出新的翻译结果，不要进行任何解释。
输入:
text: {text}
translation: {translation}
输出:
\end{verbatim}

\noindent (given the source text and its initial translation from Stage~1, the model adjusts the translation's length so that the target/source syllable ratio falls within \texttt{\{range\}}, adding or removing content as needed, and outputs the revised translation directly.)

All baselines thus receive the same contextual information and an explicit, quantitative syllable budget where applicable. The fact that even these carefully designed prompts achieve only limited success (Table~3) underscores our central finding: instruction-following alone is insufficient for rigorous syllable-level control, motivating \textsc{Homura}'s RL-based approach.